\documentclass[lettersize,journal]{IEEEtran}
\usepackage{amsmath,amsfonts}
\usepackage{array}
\usepackage[caption=false,font=normalsize,labelfont=sf,textfont=sf]{subfig}
\usepackage{textcomp}
\usepackage{stfloats}
\usepackage{url}
\usepackage{verbatim}
\usepackage{graphicx}
\usepackage{multirow}
\usepackage{amsthm}

\usepackage{bm}
\usepackage{color}
\usepackage{nomencl}
\usepackage{rotating}
\usepackage[linesnumbered,ruled,vlined]{algorithm2e}
\usepackage[hidelinks]{hyperref}

\usepackage{amsthm}
\usepackage{amssymb}
\usepackage[noend]{algpseudocode}

\usepackage{amsmath}
\usepackage{float}
\usepackage{pdflscape} 
\usepackage{adjustbox} 
\usepackage{caption} 
\usepackage{tabularx}
\usepackage{booktabs}
\usepackage{tikz}

\makenomenclature
\def\BibTeX{{\rm B\kern-.05em{\sc i\kern-.025em b}\kern-.08em
    T\kern-.1667em\lower.7ex\hbox{E}\kern-.125emX}}
\usepackage{balance}
\begin{document}
\title{Motion Planning for Robotics: A Review for Sampling-based Planners}
\author{Liding Zhang*$^{1}$, Kuanqi Cai*$^{2,3}$, Zewei Sun$^{1}$, Zhenshan Bing$^{1}$, Chaoqun Wang$^{4}$, Luis Figueredo$^{3,5}$, \\Sami Haddadin$^{3}$, ~\IEEEmembership{Fellow,~IEEE}, Alois Knoll$^{1}$, ~\IEEEmembership{Fellow,~IEEE} 

\thanks{$^{1}$L. Zhang, Z. Sun, Z. Bing and A. Knoll are with Chair of Robotics, Artificial Intelligence and Real-time Systems, TUM School of Computation, Information and Technology (CIT), Technical University of Munich, Munich, 85748, Germany. {\tt\small liding.zhang@tum.de}}
\thanks{$^{2}$K. Cai is with the Human-Robot Interfaces and Interaction(HRI2), Istituto Italiano Di Tecnologia (IIT), Genova, 16163, Italy.}
\thanks{$^{3}$K. Cai, L. Figueredo, and S. Haddadin are with the Chair of Robotics and Systems Intelligence, Munich Institute of Robotics and Machine Intelligence (MIRMI), Technical University of Munich, Munich, 80992, Germany.}
\thanks{$^{4}$C. Wang is with the School of Control Science and Engineering, Shandong University, Shandong, 250100, China.}
\thanks{$^{5}$L. Figueredo is with the School of Computer Science, University of Nottingham, Nottingham, 315199, United Kingdom.
}
\thanks{$^{*}$Equal contribution.}
}
\markboth{IEEE Transactions on \LaTeX}%
{How to Use the IEEEtran \LaTeX \ Templates}

\maketitle

\begin{abstract}
Recent advancements in robotics have transformed industries such as manufacturing, logistics, surgery, and planetary exploration. A key challenge is developing efficient motion planning algorithms that allow robots to navigate complex environments while avoiding collisions and optimizing metrics like path length, sweep area, execution time, and energy consumption. Among the available algorithms, sampling-based methods have gained the most traction in both research and industry due to their ability to handle complex environments, explore free space, and offer probabilistic completeness along with other formal guarantees. 
Despite their widespread application, significant challenges still remain. To advance future planning algorithms, it is essential to review the current state-of-the-art solutions and their limitations. In this context, this work aims to shed light on these challenges and assess the development and applicability of sampling-based methods. Furthermore, we aim to provide an in-depth analysis of the design and evaluation of ten of the most popular planners across various scenarios. Our findings highlight the strides made in sampling-based methods while underscoring persistent challenges. This work offers an overview of the important ongoing research in robotic motion planning. 
\end{abstract}


\section{Introduction}
\label{sec:introduction}
\IEEEPARstart{I}{n} recent years, robotics technology has rapidly advanced across various industries, including manufacturing~\cite{pedersen2016robot,matheson2019human,arents2022smart}, logistics~\cite{echelmeyer2008robotics,lin2021automated,bernardo2022survey}, robotic surgery~\cite{lanfranco2004robotic,diana2015robotic,Sozzi2019dynamic}, and planetary exploration~\cite{wilcox1992robotic,oberlander2014multi,Albee2020real,Chen2024}, bringing profound changes. Among the challenges in robotics, developing efficient and effective motion planning algorithms that help robots navigate complex environments, avoid obstacles, and complete tasks with minimal energy consumption and time is a critical task. The core objective of motion planning is to find the optimal path from the starting point to the target location while considering various constraints, such as dynamic environments and non-holonomic motion restrictions. This issue has become a central research topic in the field of robotics.

\begin{figure*}[h]
  \centering
  \includegraphics[width=0.92\textwidth]{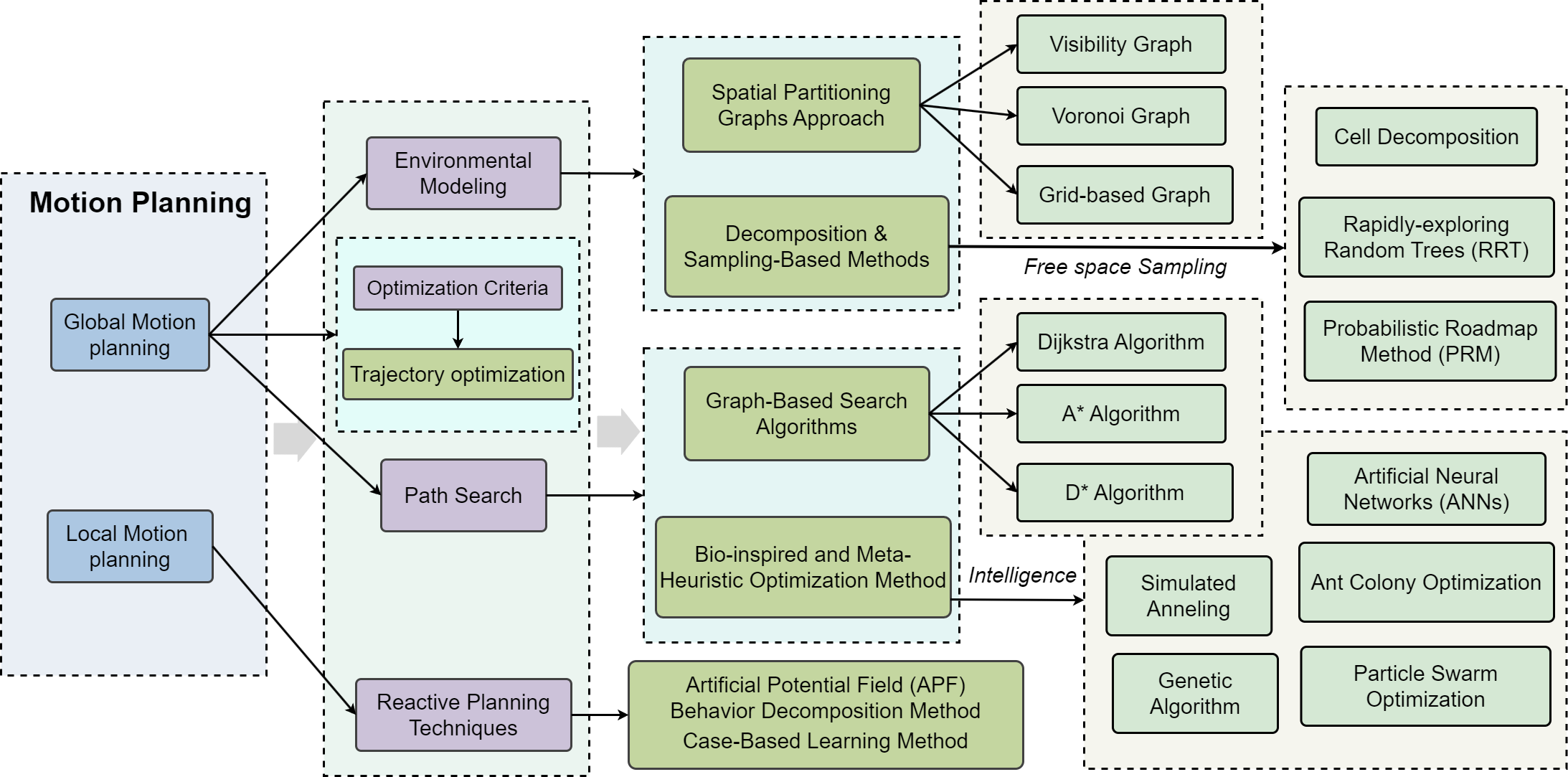}
  \caption{Overview of the most common motion planning techniques in robotics research.} \label{fig:generalPathplan}
\end{figure*}

\begin{figure}[ht] 
    \centering 
    \includegraphics[width=0.45\textwidth]{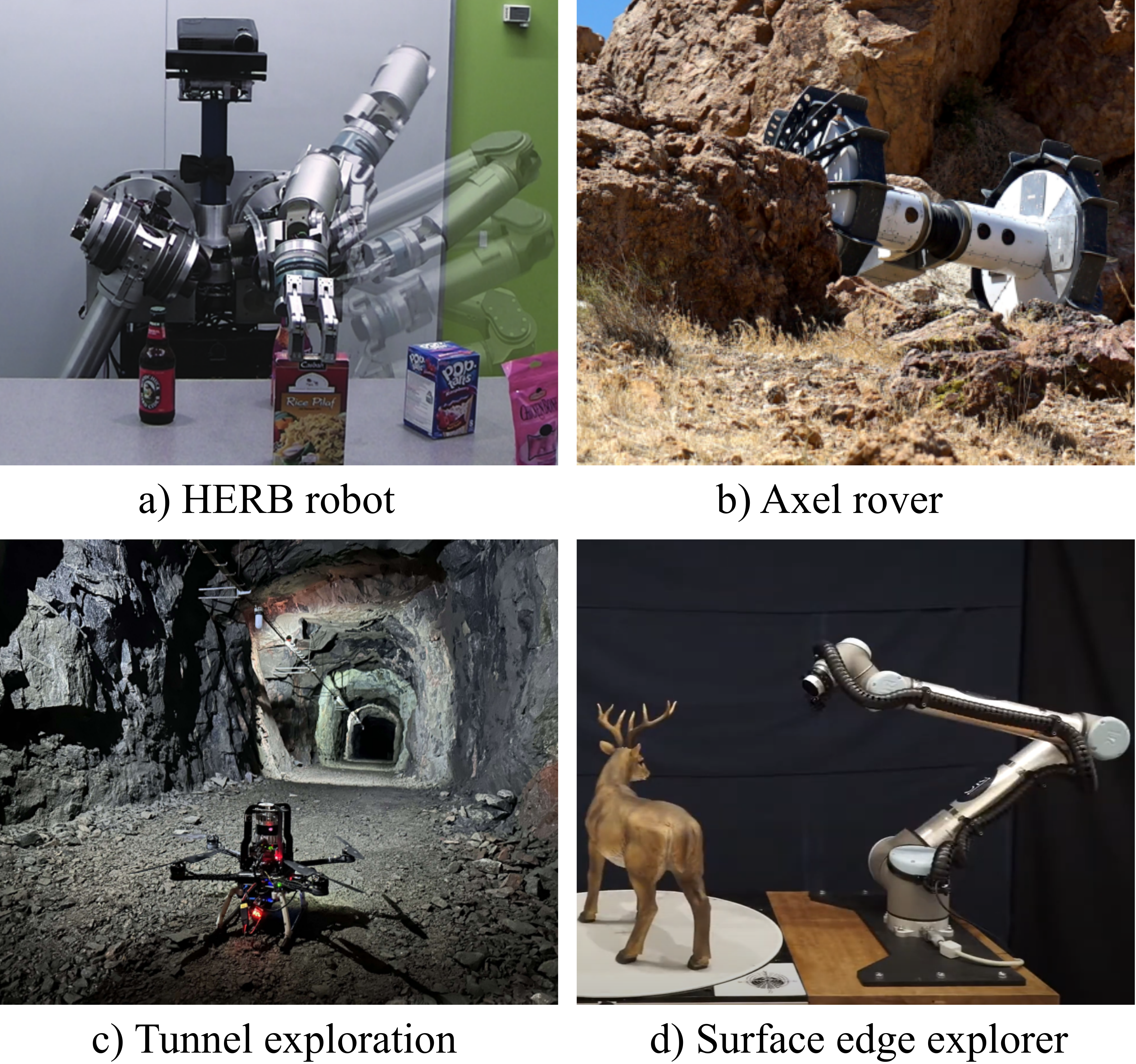} 
    \caption{Illustrates the diverse applications of sampling-based planners across multiple domains, highlighting their versatility: (a) manipulation of high-dimensional spaces by the HERB robot~\cite{Gammell2014informed}; (b) navigation of the Axel rover (UGV) over uneven terrain~\cite{strub2020adaptively}; (c) UAV exploration within constrained tunnels utilizing tree-based planning methods~\cite{Lindqvist2024}; and (d) 3D construction planning in the context of the Next Best View (NBV) problem~\cite{borderijrr24}.
} 
    \label{fig:robot_pic} 
\end{figure}

The state-of-the-art and most common algorithms related to solving the motion planning problem are shown in Fig.~\ref{fig:generalPathplan}. Motion planning can be divided into global motion planning and local motion planning. Spatial partitioning approach within configuration space~\cite{lozano1990spatial}, graph-based search algorithms, sampling-based methods, Bio-inspired and meta-heuristic optimization methods belong to global motion planning. Methods like Artificial Potential Field (APF)~\cite{vadakkepat2000evolutionary} and Dynamic Window Approach (DWA)~\cite{Fox1997} belong to local motion planning for real-time and dynamic obstacle avoidance.  Graph-based algorithms like Dijkstra~\cite{dijkstra1959note}, wavefront~\cite{zelinsky1993planning}, A*~\cite{duchovn2014path} and D*~\cite{stentz1994optimal} are resolution-complete but are computationally expensive for high dimensional complex problems. Bio-inspired and meta-heuristic optimization methods, such as Genetic Algorithm~\cite{tu2003genetic}, Particle Swarm Optimization~\cite{zhang2013robot}, and Ant Colony Optimization~\cite{garcia2009path}, are well-suited for solving multi-objective optimization problems. However, Like many other evolutionary methods, such as Simulated Annealing~\cite{zhu2006robot}, Artificial  Neural Networks~\cite{yang2004neural}, often face issues like getting stuck in local optima and having high computational costs. Additionally, they are highly sensitive to the size of the search space and the problem's data representation scheme. The APF method's advantage is its real-time adaptability for obstacle avoidance, while its drawbacks include susceptibility to local minima and difficulty in navigating between closely spaced obstacles. Among the various methods developed to address this challenge, sampling-based algorithms have emerged as a powerful and versatile approach, particularly suitable for high-dimensional and complex environments, as illustrated in Fig.~\ref{fig:robot_pic}.


Sampling-based algorithms, such as the Probabilistic Roadmap (PRM;~\cite{Kavraki1996probabilistic}), Expansive Space Trees (EST)~\cite{Hsu2002}, and Rapidly-exploring Random Trees (RRT;~\cite{lavalle1998rapidly}), have gained main attention due to their ability to efficiently handle the computational complexity associated with planning in high-dimensional spaces. These algorithms work by randomly sampling the configuration space of the robot and incrementally building a graph or tree that represents feasible paths. This approach contrasts with traditional deterministic methods, which often struggle with the curse of dimensionality and require comprehensive environment modeling. Additionally, these methods are highly adaptable and can be easily extended to accommodate various constraints and optimization criteria, making them suitable for a broad spectrum of robotic applications. The flexibility and efficiency of sampling-based approaches have led to extensive research and numerous enhancements, resulting in a rich body of literature that continues to evolve. 

In recent decades, researchers in sampling-based motion planning have published numerous comprehensive review papers. An early exploration of the field's history, documented in the review by Latombe~\cite{latombe1999motion}, discusses the advancements in motion planning, from the initial stages of basic collision detection to the current ability to handle multi-degree-of-freedom robots in complex environments, with non-robotic applications emerging as a key driver for future research. Lindemann and LaValle~\cite{lindemann2005current} review the development of sampling-based motion planning during the early 2000’s. The survey by Tsianos et al.~\cite{tsianos2007sampling} focus on the application of sampling-based motion planning in realistic settings. Elbanhawi and Simic~\cite{elbanhawi2014sampling} provide a comprehensive overview of various planners, along with a collection of common primitives applicable to a wide range of algorithms. Furthermore, Noreen et al.~\cite{noreen2016optimal} offer a detailed review of RRT* based path planning approaches, Kingston et al.~\cite{kingston2018sampling} provide an overview and discuss the problem of motion planning with constraints, while Cai et al.~\cite{cai2020mobile} focus on mobile robot motion planning in dynamic environments, Sanchez et al.~\cite{sanchez2021path} focus on autonomous mobile robots. Respectively, Gammell and Strub~\cite{gammell2021asymptotically} present an extensive overview of asymptotically optimal planners. Subsequently, Orthey et al.~\cite{orthey2023sampling} provide an excellent review of sampling-based motion planning. Recently, the paper~\cite{xu2024recent} reviews the research on RRT-based improved algorithms from 2021 to 2023, including theoretical improvements and application implementations. With the rapid advancement of sampling-based methods, conducting a survey on the current developments in sampling-based planners is highly valuable and relevant. By systematically reviewing the existing literature and identifying key trends and developments, this survey aims to serve as a valuable resource for researchers and practitioners seeking to understand and advance the capabilities of sampling-based motion planning.

This survey provides a comprehensive overview of the  sampling-based motion planning algorithms. Section~\ref{sec:problem} introduces the motion planning problem and fundamental concepts. In Section~\ref{sec:OptimalPP}, we delve into a detailed analysis of traditional sampling-based motion planning algorithms. Section~\ref{sec:limitation} explores the various applications and limitations of these algorithms. In Section~\ref{sec:experiment}, we evaluate the performance of ten popular planners across both simulated random scenarios and manipulation tasks in different dimensions. Finally, in Section~\ref{sec:conclusion} we make a conclusion.

\section{Problem Definition}
\label{sec:problem}

\subsection{Definition of Planning Problem}
Firstly, we define the path-related parameters based on the formal definition provided by Karaman and Frazzoli~\cite{Karaman2011sampling}. 
Let \( \mathcal{X} = (0, 1)^d \) represent the configuration space, where \( d \in \mathbb{N} \) and \( d \geq 2 \). Define \( \mathcal{X}_{\text{not}} \) as the obstacle region, such that \( \mathcal{X} \setminus \mathcal{X}_{\text{not}} \) is an open set. The obstacle-free space is denoted as \( \mathcal{X}_{\text{free}} = \text{cl}(\mathcal{X} \setminus \mathcal{X}_{\text{not}}) \), where \( \text{cl}(\cdot) \) denotes the closure of a set. The initial condition \( x_{\text{start}} \) is an element of \( \mathcal{X}_{\text{free}} \), and the goal region \( \mathcal{X}_{\text{goal}} \) is an open subset of \( \mathcal{X}_{\text{free}} \). A path planning problem is thus defined by the triplet \( (\mathcal{X}_{\text{free}}, x_{\text{start}}, \mathcal{X}_{\text{goal}}) \).

Consider a function \( \sigma : [0, 1] \to \mathbb{R}^d \) that represents a path in a \( d \)-dimensional configuration space. A path \( \sigma \) is considered valid if it is continuous and its total variation, denoted by \( TV(\sigma) \), is bounded. The collection of all such valid paths is denoted by \( \Sigma \).

The total variation for a path \( \sigma \), that is, \( TV(\sigma) \), is defined as the supremum of the sum of Euclidean distances between successive points along the path. That is,
\begin{equation}
TV(\sigma) = \sup_{\{n \in \mathbb{N}, 0 = \tau_0 < \tau_1 < \dots < \tau_n = s\}} \sum_{i=1}^{n} \|\sigma(\tau_i) - \sigma(\tau_{i-1})\|.
\end{equation}
When \( TV(\sigma) < \infty \), it indicates that the path \( \sigma \) has bounded variation, making it a valid path belonging to the set \( \Sigma \).

Secondly, the concept of path-planning problems in robotics and automation can be broadly classified into two primary categories:

1) {Feasible Path Planning}: This type of path planning focuses on identifying a route that effectively moves from a start point to a specified goal, ensuring that the path is achievable. For instance, the path must avoid obstacles or comply with constraints in the environment. There are often multiple possible solutions to a feasible path planning problem, but the primary objective is directly to reach the goal without concern for the path's quality or efficiency.

\noindent \textbf{Definition 1 (The Feasible Path Planning Problem)}: Given a path planning problem \( (\mathcal{X}_{\text{free}}, x_{\text{start}}, \mathcal{X}_{\text{goal}}) \), the feasible path planning problem aims to find any valid path from the start to the goal, according to a predefined rule, or to report failure if no such path exists:

\begin{equation}
\sigma \in \{\sigma_0 \in \Sigma \mid \sigma_0(0) = x_{\text{start}}, \sigma_0(1) \in \mathcal{X}_{\text{goal}}\}
\end{equation}

2) {Optimal Path Planning}: This type of path planning goes beyond simply finding a valid route by focusing on identifying the most efficient path according to specific criteria. The objective might be to minimize factors such as distance traveled, time taken, energy usage, or other relevant task-specific factors. Consequently, the optimal path planning problem seeks to determine the best possible solution among all feasible options.

\noindent \textbf{Definition 2 (The Optimal Path Planning Problem)}: Given a path planning problem \( (\mathcal{X}_{\text{free}}, x_{\text{start}}, \mathcal{X}_{\text{goal}}) \) and a cost function \( c : \Sigma \to \mathbb{R}_{\geq 0} \), the task is to find a feasible path \( \sigma^* \) such that
\begin{equation}
c(\sigma^*) = \min \{c(\sigma) : \sigma \text{ is feasible}\}.
\end{equation}
\begin{equation}
\begin{split}
\sigma^* = \arg\min_{\sigma \in \Sigma} \{c(\sigma) \mid \sigma(0) = x_{\text{start}}, \sigma(1) \in \mathcal{X}_{\text{goal}}, \\ \forall t \in [0, 1], \sigma(t) \in \mathcal{X}_{\text{free}}\}
\end{split}
\end{equation}
where \( \mathbb{R}_{\geq 0} \) represents the set of non-negative real numbers. The cost of this optimal path is denoted as \( c^* \).

\subsection{Analysis of sampling-based planning}
Probabilistic Completeness is an important property for sampling-based path
planning algorithms, which means that if there exists a feasible path from the starting position to the target position, the probability that the algorithm finds this path approaches 1 as the number of samples increases infinitely.

\noindent \textbf{Definition 3 (Probabilistic Completeness)}: An algorithm is said to be \textit{probabilistically complete} if the probability of finding a feasible path, assuming one exists, converges to 1 as the number of samples $q$ goes to infinity. Mathematically, this is represented as:
\begin{equation}
\liminf_{q \to \infty} P(\Sigma_q \neq \emptyset) = 1,
\end{equation}
where $\Sigma_q$ is the set of feasible paths found using $q$ samples, and $P$ denotes probability

\noindent \textbf{Definition 4 (Asymptotic Optimality Definition)}:
An algorithm is said to be asymptotically optimal if, for any positive number $\epsilon$, the following holds:
\begin{equation}
\lim_{q \to \infty} P(c(\pi_q) - c^* < \epsilon) = 1,
\end{equation}
where $c(\pi_q)$ is the cost of the path found with $q$ samples, $c^*$ is the cost of the optimal path, and $P$ denotes probability.

\section{Optimal Path Planning with Sampling-based Methods}
\label{sec:OptimalPP}
\subsection{Sampling-based Planner Framework}

To find the optimal path in the planning task, the Rapidly-exploring Random Tree star-based planner (RRT*)~\cite{Karaman2011sampling} is widely used. The workflow of RRT* is shown in Algorithm 1. First, the search tree $\mathcal{T}$ is initialized with the root node $x_{\text{start}}$. Random sample points $x_{\text{rand}}$ are generated through the \textbf{\textit{Sampling}}($\cdot$) function within the search space. Then, the \textbf{\textit{Nearest}}($\cdot$) function identifies the nearest node $x_{\text{near}}$ in the tree $\mathcal{T}$ to the sample point $x_{\text{rand}}$. Next, the \textbf{\textit{Steer}}($\cdot$) function moves from $x_{\text{near}}$ towards $x_{\text{rand}}$ by a defined step size to obtain a new node $x_{\text{new}}$ and creates the edge $(x_{\text{near}}, x_{\text{new}})$. After that, the \textbf{\textit{CollisionFree}}($\cdot$) method checks whether this edge collides with any obstacles or boundaries in the map. If no collision is detected, the tree is expanded by adding $x_{\text{new}}$ to the set of vertices and the edge $(x_{\text{near}}, x_{\text{new}})$ to the set of edges. If a collision is detected, the newly generated node is discarded, and the algorithm retries the sampling process. Once $x_{\text{new}}$ is added, the algorithm performs an \textbf{\textit{Extend}}($\cdot$) operation by searching within a defined neighborhood around $x_{\text{new}}$ to identify a parent vertex that minimizes the overall cost from the initial state. A \textbf{\textit{Heuristic}}($\cdot$) method is employed at this stage to estimate the cost-to-go from $x_{\text{new}}$ to the goal, allowing the algorithm to prioritize more promising nodes and improve convergence towards the optimal solution. 
Subsequently, a \textbf{\textit{Rewire}}($\cdot$) step is conducted by checking neighboring vertices to see if their connection through $x_{\text{new}}$ reduces their path cost. If so, their parent vertices are updated accordingly to reflect the more efficient path.

As the number of iterations increases, the sampling-based algorithm progressively generates paths that approach the optimal solution. Ultimately, the algorithm outputs a path that is refined with each iteration, ensuring it is close to the shortest possible path.

\begin{algorithm}[t!]
\caption{Rapidly Exploring Random Tree Star (RRT*)}

\DontPrintSemicolon
\SetKwInOut{Input}{Input} 
\SetKwInOut{Output}{Output} 
\SetKwFunction{initializeTree}{initializeTree}
\SetKwFunction{insertNode}{insertNode}
\SetKwFunction{sampling}{sampling}
\SetKwFunction{nearest}{nearest}
\SetKwFunction{collisionFree}{collisionFree}
\SetKwFunction{extend}{extend}
\SetKwFunction{steer}{steer}
\SetKwFunction{chooseParent}{chooseParent}
\SetKwFunction{cost}{cost}
\SetKwFunction{near}{near}
\SetKwFunction{rewire}{rewire}

\Input{Start state $x_\textit{start}$, goal region $\mathcal{X}_{\text{goal}}$}
\Output{Feasible tree $\mathcal{T}$}

$V \gets x_{start}$, $E \gets \emptyset$, $\mathcal{T} = (V, E) \gets RRT^*(x_{start})$

\For{$i \gets 1$ \KwTo $N$}{
    $x_{rand} \gets \sampling(i)$\;
    $x_{nearest} \gets \nearest(\mathcal{T}, x_{rand})$\;
    $x_{new} \gets \steer(x_{nearest}, x_{rand})$\;
    \If{\collisionFree($x_{nearest}$, $x_{new}$)}{
        $X_{near} \gets \near(\mathcal{T}, x_{new}, k$ \text{or} $r)$\;
        $x_{min} \gets \arg\min\limits_{x_{near} \in X_{near}} \left( \cost(x_{near}) + \|x_{near} - x_{new}\| \right)$\\\Comment{heuristic for minimal path cost, $\| \cdot \|$ denote as $L^2$ norm}\\
        $\mathcal{T} \gets \extend(\mathcal{T}, x_{new})$\;
        $\rewire(\mathcal{T}, X_{near}, x_{min}, x_{new})$\;

        \If{$x_{new} \in \mathcal{X}_{\text{goal}}$}{
        \Return{$\mathcal{T}$}\;
        }
    }
}
\Return{\text{failure}}\;

\end{algorithm}

\begin{figure*}[t!] 
    \centering 
    \includegraphics[width=0.99\textwidth]{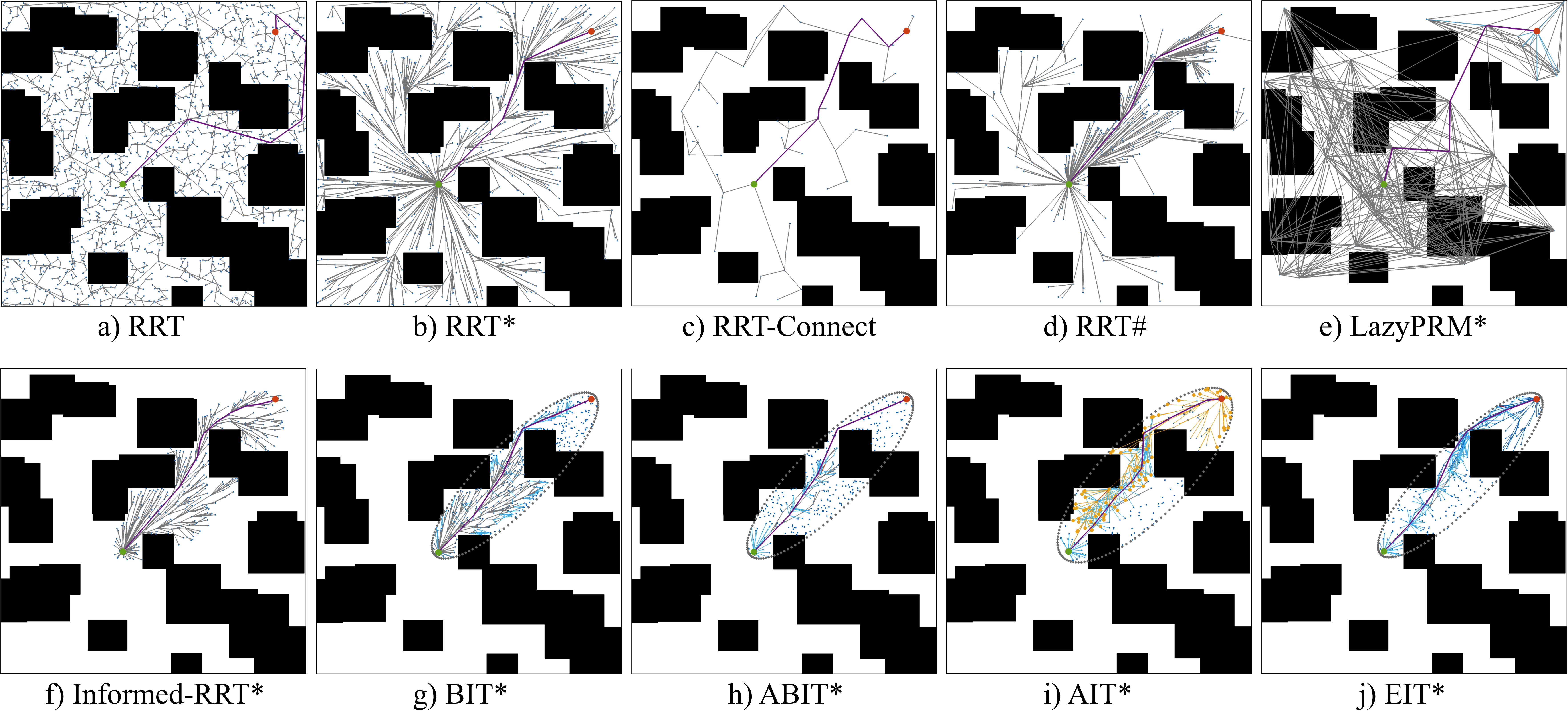} 
    \caption{The 2D representation of the of the search trees constructed by RRT, RRT$^*$, RRTConnect, RRT$^\#$, lazyPRM$^*$, Informed RRT$^*$, BIT$^*$, ABIT$^*$, AIT$^*$, and EIT$^*$ to find an initial solution when optimizing path length (a–j). The start and goal are represented by a green dot and a red dot, respectively. Sampled states are shown as small black dots. Obstacles in the state space are indicated by black rectangles. The initial solutions are depicted in purple. (a) RRT randomly explores the state space, fully evaluating many unnecessary edges for both objectives. (b) RRT$^*$ similarly evaluates many edges that don't contribute to the initial solution for both path length and clearance. (c) RRT-Connect is more efficient, connecting start and goal faster, but still evaluates redundant edges. (d) RRT$^\#$ reduces redundant edge evaluation using graph refinement. (e) lazyPRM$^*$ avoids full evaluation by deferring edge checking until necessary. (f) Informed RRT$^*$ refines its exploration by focusing on promising regions. (g) BIT$^*$ ranks edges by potential solution cost, reducing unnecessary evaluations. (h) ABIT$^*$ improves BIT$^*$ by incorporating more efficient search heuristics. (i) AIT$^*$ utilizes problem-specific heuristics to further reduce redundant evaluations. (j) EIT$^*$ combines cost and effort heuristics, making it robust even when problem-specific heuristics are unavailable.
} 
    \label{fig:planner_compare} 
\end{figure*}
Researchers have proposed various strategies to improve different components of the sampling-based planner framework, including the $\textbf{\textit{Sampling}}(\cdot)$ function, $\textbf{\textit{Nearest}}(\cdot)$ function, $\textbf{\textit{Steer}}(\cdot)$ function, $\textbf{\textit{CollisionFree}}(\cdot)$ method, $\textbf{\textit{Extend}}(\cdot)$ operation, \textbf{\textit{Heuristic}}($\cdot$) method and $\textbf{\textit{Rewire}}(\cdot)$ step. Fig.~\ref{fig:planner_compare} provides a general overview of these technological advancements, which will be further detailed in the following sections.

\subsection{Advanced Sampling Approaches}\label{sec:Advanced}
\begin{figure*}[t!]
    \centering
    \begin{tikzpicture}
    \node[anchor=center] at (0,0) 
    {\includegraphics[width=0.8\textwidth]{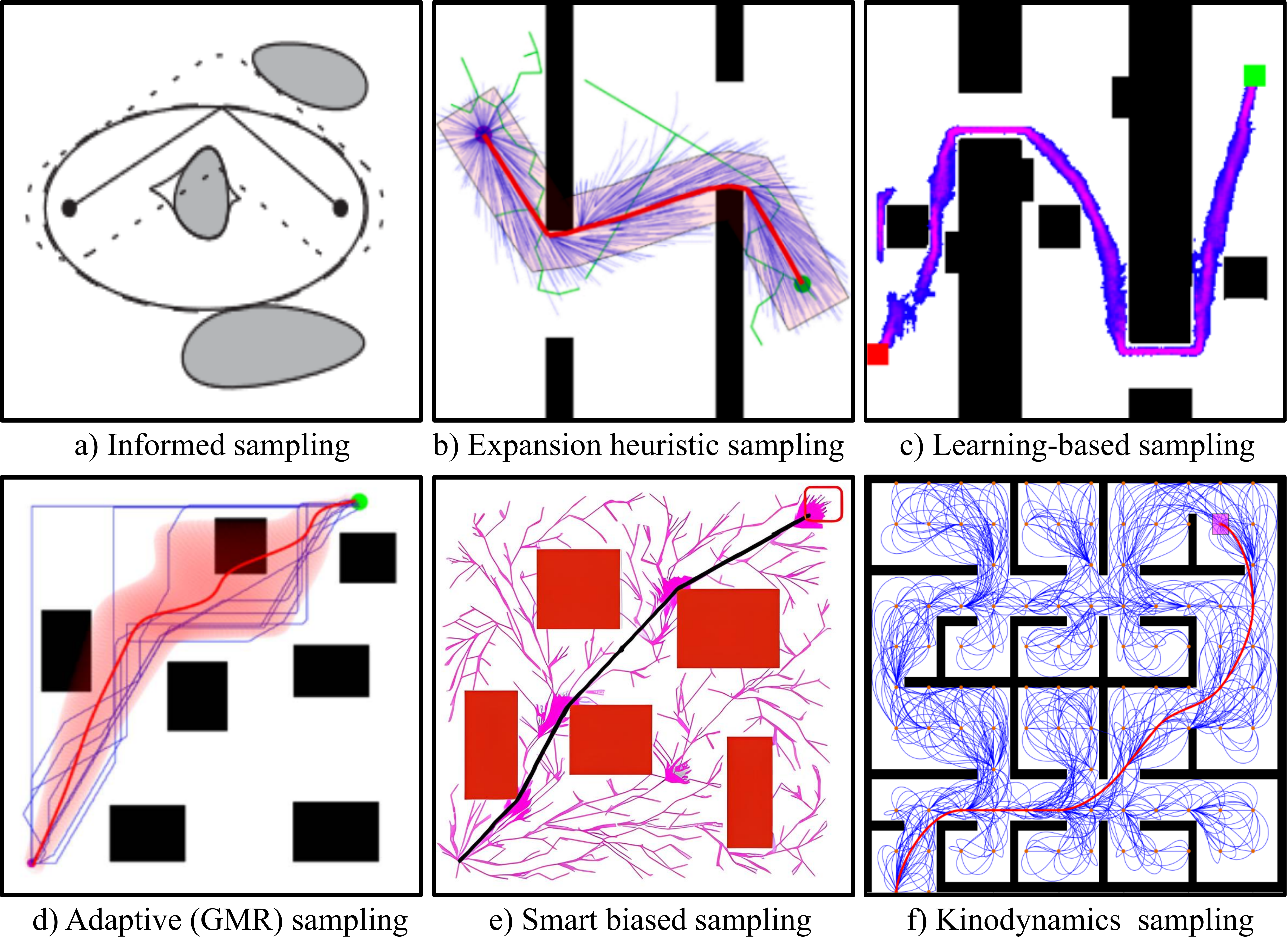}};
    \end{tikzpicture}
    \caption{(a): Informed RRT*~\cite{Gammell2014informed,gammell2018informed} enhances the standard RRT* by focusing sampling in an ellipsoidal region that contains the optimal path, thereby improving convergence speed and solution quality. (b): EP-RRT*~\cite{ding2023improved} expands the initial path to define an "expansion area," where heuristic sampling is performed to refine the path and converge to an optimal solution more efficiently. (c): Neural-RRT*~\cite{Wang2020neural} combines RRT* with learning-based features (e.g., neural networks) and optimizations for sampling prediction and convergence time reduction, resulting in rapid pathfinding in cluttered environments. (d): GMR-RRT*~\cite{Wang2022} incorporates demonstrated trajectories into a Gaussian Mixture Model to define exploration-worthy regions. (e): RRT*-Smart~\cite{Nasir2013RRT*-SMART} utilizes the triangular inequality to improve paths and employs intelligent sampling techniques for the path edges. (f): This method~\cite{sakcak2019sampling} presents pre-computed kinodynamic motion primitives for sampling and efficient path planning.}
    \label{fig:sampling}
    \vspace{0.5em}
\end{figure*}
The \textbf{\textit{Sampling}}($\cdot$) function generates an infinite sequence of sample points within the search space. Sampling functions are classified as 
 \textbf{Informed Set},  \textbf{Learning-based}, \textbf{Adaptive Sampling}, \textbf{Smart Biased Sampling} and \textbf{Kinodynamic}.

\textbf{Informed Set}, such as Informed RRT*~\cite{Gammell2014informed,gammell2018informed}(Fig.~\ref{fig:sampling}, a), which restricts the node sampling range by creating a hyper-ellipsoid subset~\cite{Zhang2023} for sampling and conducts direct sampling within this subset. By constraining the algorithm's sampling region, this subset increases the probability of selecting relevant nodes, thereby boosting the algorithm's efficiency. BIT*~\cite{gammell2015batch,gammell2020batch} builds on Informed RRT*, with batch sampling in the hyper-ellipsoid subset, this leads to a further enhancement in the algorithm's efficiency. Respectively, Ding et al. ~\cite{ding2023improved} propose the expanding path RRT* (EP-RRT*)(Fig.~\ref{fig:sampling}, b) based on heuristic sampling in the path expansion area. Another paper~\cite{yu2022cyl} proposes a cylinder-based informed rapid exploration random tree (Cyl-iRRT*) path planning algorithm, which biases the sampling of new candidate states into an admissible cylindrical subset around the centerline from start to goal positions via rejection sampling to quickly find the homotopy optimal path in a 3-D environment. 

\textbf{Learning-based Sampling}, such as Neural RRT*~\cite{Wang2020neural}(Fig.~\ref{fig:sampling}, c) and Neural Informed RRT*~\cite{huang2024neural},  guides the sampling direction by training neural networks, which are particularly effective in navigating narrow passages and complex environments. Another paper ~\cite{ichter2018learning} uses a conditional variational auto-encoder (CVAE) to learn sampling distributions, Molina et al.~\cite{molina2020learn} propose to use convolutional neural networks (CNNs) to identify critical regions for robot planning so that the sampling process can be biased to these regions.

\textbf{Adaptive Sampling}, as demonstrated in methods like the Rapidly-Exploring Adaptive Sampling Tree* (RAST*)~\cite{xiong2020rapidly}, Gaussian Mixture Regression 
Rapidly exploring Random Tree* (GMR-RRT*)~\cite{Wang2022}(Fig.~\ref{fig:sampling}, d) and the similar works such as\cite{cai2021risk,cai2022curiosity}, leverages an adaptive sampling approach to prioritize information-dense regions. This strategy effectively reduces localization uncertainty by concentrating on areas with higher information gain.

\textbf{Smart Biased Sampling}, like RRT*-SMART~\cite{Nasir2013RRT*-SMART}(Fig.~\ref{fig:sampling}, e), mRRT*-Smart~\cite{tak2019improvement} and RRT*SMART-A*~\cite{suwoyo2023integrated} employs intelligent sampling around key nodes from an initial path, accelerating convergence. 

\textbf{Kinodynamics}, Boeuf et al.~\cite{boeuf2015enhancing} present an incremental state-space sampling technique to avoid generating local trajectories that violate kinodynamic constraints. Other papers prioritize generating samples that maximize clearance, meaning the distance between the robot and its surroundings. This is typically done by sampling a feasible state and taking random steps to further increase the clearance~\cite{sakcak2019sampling,verginis2022kdf}(Fig.~\ref{fig:sampling} f).

\subsection{Nearest Neighbor Methods}\label{sec:nearest}
\begin{figure}[t]
    \centering
    \begin{tikzpicture}
    \node[anchor=center] at (0,0) 
    {\includegraphics[width=0.48\textwidth]{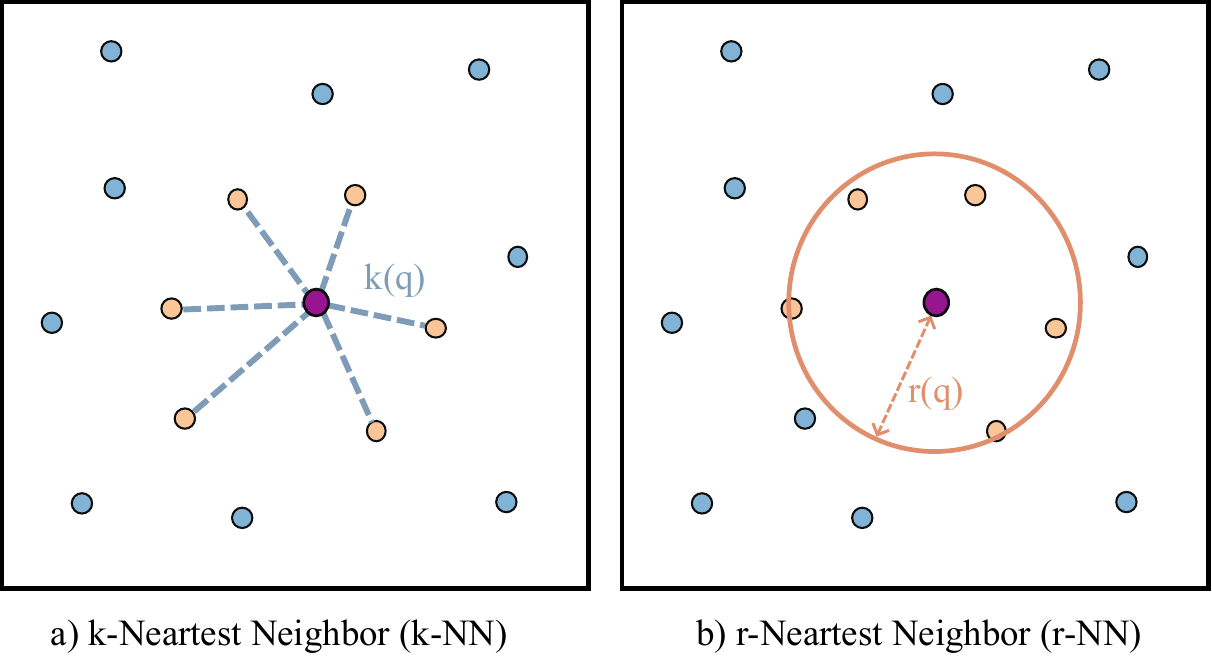}};
    \end{tikzpicture}
    \caption{The figure illustrates two nearest neighbor strategies. Panel (a) illustrates the $k$-nearest neighbors approach, which uses a constant number of neighbors $k(q)$ as defined in Eq.~\ref{eq:knn}. Panel (b) depicts the $r$-nearest neighbors strategy, where the radius $r(q)$, as defined in Eq.~\ref{eq:rnn}, dynamically changes based on the current number of samples per batch.
}
    \label{fig:knnRNN}
    \vspace{0.5em}
\end{figure}
The \textbf{\textit{Nearest}}($\cdot$) function refers to a key step in the sampling-based algorithms where the nearest node in the tree is found for a randomly sampled point and used as the base to expand the tree towards that point.

Traditional nearest-neighbor searches are computationally expensive due to their linear search nature. Techniques like KD trees~\cite{Atramentov2002efficient} address these challenges by employing advanced data structures and algorithms. \textbf{KD trees} partition the space recursively, enabling efficient multi-dimensional searches. This method improves the search process compared to a linear search, with an average time complexity of logarithmic order, making it highly suitable for large, high-dimensional datasets.

In contrast, the \textbf{k-nearest neighbor (k-NN) strategy} in sampling-based algorithms can also be implemented using formulas that adaptively determine the number of neighbors \(k\) for each node based on the number of samples \(q\) and the dimensionality of the space \(n\). The number of neighbors \(k(q)\) is calculated as:
\begin{equation}
\label{eq:knn}
    k(q) := \eta e^{\left(1 + \frac{1}{n}\right) \log(q)},
\end{equation}
where \(\eta > 1\) is a tuning parameter. This strategy ensures that the tree expands efficiently while maintaining good coverage of the space. Pan et al.~\cite{pan2013faster} propose a novel approach for rapid probabilistic collision checking aimed at boosting the efficiency of sampling-based motion planning. In this method, the k-NN strategy is employed to identify the closest prior query sample to the new query configuration. The results demonstrate that this integrated approach enhances the RRT-based path planner by speeding up local pathfinding and optimizing the search sequence on the roadmap. These findings have been validated on both rigid and articulated robots.


Respectively, the \textbf{r-nearest neighbor (r-NN) strategy} used in the r-disc (i.e., \(r(q)\) as connection radius) to connect nearest nodes is computed by:

\begin{equation}
\label{eq:rnn}
r(q) := 2\eta \left(1 + \frac{1}{n}\right)^{\frac{1}{n}} \left( \frac{\min \{\lambda(\mathcal{X}), \lambda(\hat{\mathcal{X}})\}}{\lambda(B_{1,n})} \right)^{\frac{1}{n}} \left( \frac{\log(q)}{q} \right)^{\frac{1}{n}},
\end{equation}
where \(\lambda(\cdot)\) denotes the Lebesgue measure, and $B_{1,n}$ is the $n$-dimensional unit ball, $\hat{\mathcal{X}}$ is defined as the informed set, this providing a way to scale the connection radius according to the sample distribution and the space's properties.

In addition to the aforementioned r-disc strategy, Kleinbort et al.~\cite{kleinbort2020collision,kleinbort2020refined} demonstrate that the r-disc approach can achieve superior performance compared to the k-nearest variant, though the connection radius \(r(q)\) must be calibrated according to the characteristics of the state space. Properly tuning the connection radius for specific applications can improve search efficiency. Furthermore, faster-decreasing radii are presented by Janson et al.~\cite{janson2015fast,janson2018deterministic} and Tsao et al.~\cite{tsao2020sample}, enabling the r-disc strategy to perform even better when handling varying dimensions and sampling densities.

\subsection{Tree-based Expansion Strategies}\label{sec:Expansion}
Based on the step size constraint, from the nearest node, the \textbf{\textit{Steer}}($\cdot$) function takes a small step in the direction of the random point, generating a new node, and creates the edge. Wang et al.~\cite{wang2016variant,wang2017autonomous} present variant step size RRT, which adaptively changes the step size of the tree according to the location of obstacles. Similarly, another paper~\cite{zhang2021improved} changes the step size adaptively  according to the density of obstacles. Subsequently, Yang et al.~\cite{yang2023variable} present a novel Variable Step Size (VSS) strategy based on RRT*. The VSS strategy adapts the expansion step size dynamically by considering both the direction of the vertex and the target point within the random tree, with the goal of accelerating the approach toward the target point. In another paper~\cite{shen2023adaptive} the gradient descent method is employed to adapt the step size, causing it to gradually decrease as the goal region is approached. Additionally, Li et al.~\cite{li2021path} present a gravity adaptive step size strategy.

\begin{figure}[h]
    \centering
    \begin{tikzpicture}
    \node[anchor=center] at (0,0) 
    {\includegraphics[width=0.48\textwidth]{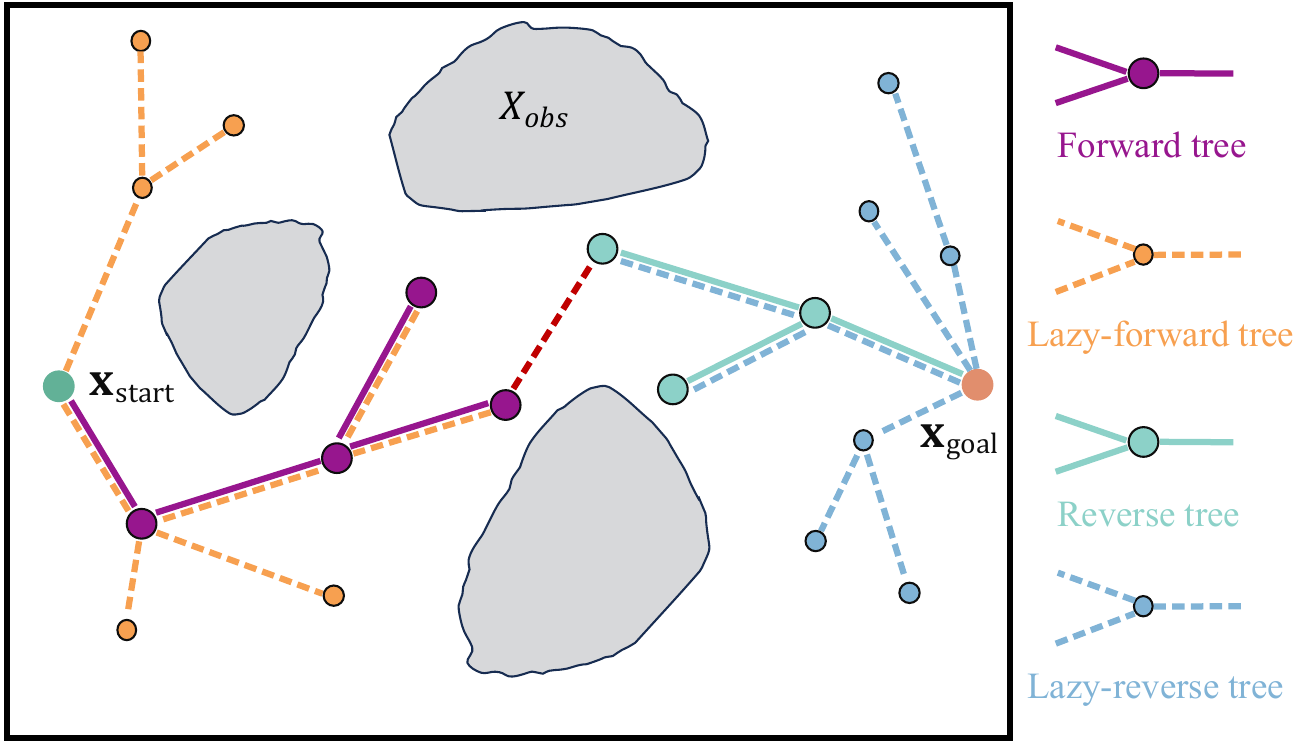}};
    \end{tikzpicture}
    \caption{The figure illustrates bidirectional search from both the start and goal states, showing four trees: the forward tree in purple, the lazy-forward tree in orange, the reverse tree in light green, and the lazy-reverse tree in blue. Bidirectional search reduces pathfinding time, while the lazy search in both the forward and reverse trees performs sparse collision checking, postponing the computationally expensive full collision check. Further details will be provided in Section~\ref{sec:collision}.
}
    \label{fig:bitrees}
\end{figure}

To quickly find a feasible path, the bidirectional expansion strategy for steering was developed. Bidirectional RRT, where two trees grow simultaneously from the start and goal points, seeking to connect with each other, as illustrated in Fig.~\ref{fig:bitrees}.

Initial approach, like RRT-Connect is introduced by Kuffner and LaValle~\cite{Kuffner2000rrt}, establishes the foundational method of constructing two trees from start and goal points and connecting them, thereby greatly improving search efficiency. However, RRT-connect is not asymptotically optimal like
RRT*. To solve this, Akgun et al.~\cite{akgun2011sampling} present Bidirectional RRT* (B-RRT*), incorporate an existing bi-directional approach to search which decreases the time to find an initial path. Similarly, Jorden et al.~\cite{jordan2013optimal} present Optimal B-RRT*, an asymptotically optimal bidirectional method that achieves an improved convergence rate by incorporating several heuristic techniques. While Qureshi et al.~\cite{qureshi2015intelligent} introduce a new variant called Intelligent Bidirectional-RRT* (IB-RRT*) which is an improved variant of the optimal RRT* and bidirectional version of RRT* (B-RRT*) algorithms. IB-RRT* utilizes the bidirectional trees approach and introduces an intelligent sample insertion heuristic for fast convergence to the optimal path solution using uniform sampling heuristics. Furthermore, Tahir et al.~\cite{tahir2018potentially} propose the Potentially Guided Intelligent Bi-directional RRT* (PIB-RRT*) and Potentially Guided Bi-directional RRT* (PB-RRT*), which are extensions of Bi-directional RRT* (B-RRT*) and Intelligent Bi-directional RRT* (IB-RRT*). These methods greatly enhance the convergence rate and utilize memory more efficiently in cluttered environments. Additionally, the B{I$^2$}RRT* algorithm~\cite{burget2016bi} builds upon the Informed RRT*~\cite{Gammell2014informed} by implementing a bidirectional search strategy, which helps in finding initial solutions faster and allows more time for refinement. Subsequently, Yi et al.~\cite{Yi2016homotopy} combine homotopy topological spaces with bidirectional trees in the Homotopy-aware RRT*, increasing the probability of finding optimal paths by exploring different topological spaces. Moreover, Lin et al.~\cite{lin2020bidirectional} propose the Bidirectional Homotopy-Guided RRT (BH-RRT), which uses obstacle contour information to guide tree growth, improving success rates. Wang et al.~\cite{wang2021efficient} introduced the Bidirectional-Unidirectional RRT Extend Function, which switches from bidirectional to unidirectional search to overcome complex boundary problems, enhancing search efficiency. Another paper~\cite{liu2019goal} present a goal-biased bidirectional RRT with curve smoothing, connecting tree parts with Bezier curves to meet kinematic constraints, thus achieving higher success rates and shorter search times. Recently, Peng et al.~\cite{xin2023improved} develop the Improved Bidirectional RRT*, utilizing an artificial potential field to reduce randomness and inflection points, resulting in shorter and more efficient paths.

\subsection{Collision Check Methods}\label{sec:collision}

After the \textbf{\textit{Steer}}($\cdot$) function creates the edge, the \textbf{\textit{CollisionFree}}($\cdot$) method involves verifying whether the path between two nodes intersects with any obstacles. This basic approach, as detailed in LaValle and Kuffner's 2001 seminal paper on Rapidly-exploring Random Trees (RRT)~\cite{lavalle2001rapidly}, involves checking the line segment between the current node and the new node for collisions. Over time, various methods have been developed to enhance the efficiency of collision checking.

Lazy Collision Checking, such as Lazy PRM~\cite{Bohlin2000path} and Fuzzy PRM~\cite{nielsen2000two}, both initially assume the path is collision-free and perform collision checks only when necessary. If a collision is later detected, the path is re-evaluated and corrected. However, neither Lazy PRM nor Fuzzy PRM guarantees solution quality. To solve this, Hauser et al.~\cite{hauser2015lazy} present two novel motion planners, Lazy-PRM* and Lazy-RRG*, they are almost-surely asymptotically optimal algorithms that grow a network of feasible vertices connected by edges. Edges are not immediately checked for collision, but rather are checked only when a better path to the goal is found. This strategy avoids checking the vast majority of edges that have no chance of being on an optimal path. Another paper~\cite{kim2018adaptive}, Adaptive Lazy Collision Checking delays collision checking in regions likely to be free of obstacles while checking early in other regions to reduce optimistic thrashing, enhancing the planner's performance in complex environments. Most recently, Neural Network Collision Checking has been introduced by Kew et al.~\cite{chase2020neural} through ClearanceNet, a neural network-based heuristic that predicts collision clearance, facilitating parallel RRT processing and significantly accelerating the collision checking process.

\subsection{Tree Extension Techniques}\label{sec:Extension}
After $x_{\text{new}}$ is added, the algorithm executes an \textbf{\textit{Extend}}($\cdot$) operation to determine the parent node of $x_{\text{new}}$, the node that results in the lowest cost from the starting point to $x_{\text{new}}$ is selected as its parent node.

LaValle and Kuffner's paper "Rapidly-exploring Random Trees: Progress and Prospects"~\cite{lavalle2001rapidly} introduce the basic principles of the standard RRT algorithm and the Extend Tree process, laying the foundation for subsequent advancements. FMT*~\cite{janson2015fastmarchingtreefast} (Fast Marching Tree) leverages the Fast Marching Method (FMM) to enhance the efficiency of RRT in high-dimensional spaces by integrating efficient sampling and searching strategies. Specifically, the FMT* algorithm uses FMM to quickly evaluate the quality of nodes when building the tree structure, accelerating the convergence of the path planning process. This makes FMT* particularly effective in complex, high-dimensional path planning problems. Subsequently, Qureshi et al.~\cite{Qureshi2016potential} introduce the P-RRT*, which integrates Artificial Potential Fields (APF) into the RRT* framework to provide a more directed exploration and faster convergence. Another strategy is Quick-RRT*~\cite{jeong2019quick}, which uses triangular inequality for parent node selection. Extending these approaches, Li et al. ~\cite{li2020pq} propose the PQ-RRT*, which combines the strengths of P-RRT* and Quick-RRT*, further enhancing the convergence speed by improving both the sampling strategy and the optimization procedures used in the tree expansion process. Moreover, Liao et al.~\cite{liao2021fff} present F-RRT*, which improves path cost by generating a parent node for the random point rather than choosing one from the existing vertices. Similarly, another paper~\cite{li2022fast} introduces Fast-RRT*. In order to achieve a path with a lower cost compared to the RRT* algorithm, the ancestors of the nearest node are taken into account up to the initial state when selecting a parent for the new node. Next, Armstrong et al.~\cite{Armstrong2021amrrt} introduce the Assisting Metric RRT* (AM-RRT*). This algorithm incorporates an assisting metric that combines Euclidean distance with a novel metric to optimize tree growth, extension, and rewiring, thereby improving coverage and path quality. This approach not only considers distance but also factors such as path smoothness or traversability, resulting in higher quality paths while maintaining efficient search performance. Respectively, Wanga et al.~\cite{wanga2021improved} present an evaluation function for growth points based on the adaptive resolution octree map, which guides the generation of RRT paths towards a more intentional extension strategy. Additionally, the algorithm reselects parent nodes and candidate nodes, followed by a rewiring process. As a result, the improved RRT algorithm eliminates redundant bifurcations in the growth tree, reduces the number of sampling instances, and significantly improves growth efficiency.

\subsection{Heuristic-guided Exploration Methods}\label{sec:Heuristic}

While the tree is extending, the \textbf{\textit{Heuristic}}($\cdot$) method is used to estimate the remaining cost from $x_{\text{new}}$ to the goal, enabling the algorithm to prioritize nodes that appear more promising and accelerate convergence toward an optimal solution.

Heuristic-guided sampling enhances the efficiency of path planning algorithms by biasing the extending process towards regions more likely to yield high-quality paths~\cite{rickert2008balancing}. The Heuristically Guided RRT (hRRT) algorithm~\cite{urmson2003approaches} proposed by Urmson and Simmons utilizes a heuristic cost function to evaluate the quality of different states, thereby optimizing the sampling process. This
algorithm has improved performance but does not provide any guarantee on the quality of its solution. To solve this, Gammell et al.~\cite{Gammell2014informed,gammell2018informed} present Informed RRT* enhances RRT* by incorporating an admissible cost heuristic, ensuring that only states capable of improving the current solution are considered. This approach accelerates the convergence rate of RRT* while preserving its asymptotic optimality with high probability. Following this, Gammell et al.~\cite{gammell2015batch,gammell2020batch} develop the Batch Informed Trees (BIT*), which compactly groups states into an implicit random geometric graph (RGG)\cite{penrose2003random}, employing step-wise search similar to Lifelong Planning A* (LPA*)\cite{koenig2004lifelong}, based on expected solution quality. This approach combines heuristic-guided search with batch sampling and heuristic sorting of the search process, ensuring rapid convergence to high-quality solutions in both low- and high-dimensional spaces. Next, the Adaptively Informed Trees (AIT*)~\cite{strub2020adaptively}, improves upon BIT* by utilizing the same progressively denser RGG approximation, but it employs an asymmetric bidirectional search. This search calculates and leverages a more accurate cost heuristic tailored to each specific RGG approximation. Then, the Advanced BIT* (ABIT*) ~\cite{strub2020advanced} further improves performance by integrating advanced graph-search techniques, such as heuristic inflation and search truncation. Furthermore, the Greedy BIT* (GBIT*)~\cite{chen2021greedy} extends ABIT* by incorporating a greedy search policy inspired by RRT-Connect~\cite{Kuffner2000rrt}, which accelerates the discovery of initial solutions and enhances convergence speed, while maintaining the asymptotic optimality of the solution. Respectively, Effort Informed Trees (EIT*)~\cite{strub2022adaptively} builds upon AIT* by utilizing problem-specific information in a way that takes advantage of informative admissible cost heuristics when available, but still performs effectively in their absence. It achieves this by incorporating additional types of problem-specific data, such as the computational effort required to validate a path. This generalization extends asymptotically optimal informed path planning algorithms to a wider range of problems, including those without effective a priori cost heuristics. Moreover, Hartmann et al.~\cite{hartmann2022effort} present Effort Informed Roadmaps (EIRM*), EIRM* extends the EIT* approach to efficiently handle multiquery problems by actively reusing computational effort and managing graph size over multiple planning queries, thereby improving both speed and efficiency in complex environments. Subsequently, Li et al.~\cite{Li2023biait} propose Symmetrical Bidirectional Optimal Path Planning with Adaptive Heuristic (BiAIT*), in contrast to AIT*, BiAIT* utilizes a symmetric bidirectional search for both heuristic calculation and space exploration. This method allows BiAIT* to find an initial solution faster than AIT* and to update the heuristic more efficiently when a collision occurs.




\subsection{Various Rewiring for Tree Optimization}\label{Rewiring}

Subsequently, a \textbf{\textit{Rewire}}($\cdot$) step is performed by evaluating neighboring vertices to determine if connecting through $x_{\text{new}}$ would lower their overall path cost. If a more efficient route is found, the parent vertices of those neighboring nodes are updated accordingly to reflect the improved path.

Karaman and Frazzoli~\cite{Karaman2011sampling} are the first to present the idea of using the rewiring method to ensure asymptotic optimality. Building on top of the RRT* presented in~\cite{Karaman2011sampling}, RRT-sharp (RRT$^\#$)~\cite{arslan2013use,Arslan2015} employs a more efficient rewiring cascade that not only propagates reduced cost-to-goal information throughout the graph but also adjusts connections within local neighborhoods when even lower cost-to-goal values are achievable. Moreover, RRT$^X$~\cite{Otte2016rrtx} introduces a rewiring strategy to adapt to changes in dynamic environments. This algorithm quickly recalculates paths when the environment changes for dynamic obstacles. Subsequently, Quick-RRT*~\cite{jeong2019quick} enhances the rewiring procedure by incorporating the ancestry of nearby vertices, instead of only checking if a new node can provide a better connection for its direct neighbors. This improves the path optimization process by generating straighter paths and reducing detours. Respectively, The Real-Time RRT* (RT-RRT*) algorithm~\cite{naderi2015rt}, based on RRT*~\cite{Karaman2011sampling} and Informed RRT*~\cite{Gammell2014informed}, introduces an online tree rewiring strategy for real-time path planning in dynamic environments. This method enables the tree to quickly adapt to changes, maintaining near-optimal paths without rebuilding the entire tree. Additionally, a triangular inequality-based rewiring method~\cite{kang2021improved} is applied to the RRT-Connect~\cite{Kuffner2000rrt} algorithm, demonstrating faster planning times and shorter path lengths compared to both RRT and RRT-Connect algorithms. Similarly, the Post Triangular Rewiring method~\cite{kang2021post} further improves the optimality of paths generated by the RRT algorithm by leveraging the triangular inequality principle. 


\subsection{Hybrid Search Approaches}\label{Hybrid}
In sampling-based algorithms, hybrid methods combine multiple path planning strategies to enhance the algorithm's performance and adaptability in complex environments.

Wei and Liu~\cite{wei2011hybridizing} combine RRT with a variable-length genetic algorithm, enhancing smooth path generation by minimizing path length and curvature, maintaining diversity, and preventing premature convergence. Mashayekhi et al.~\cite{mashayekhi2020hybrid} propose Hybrid RRT, Hybrid RRT utilizes a dual-tree search, enabling it to find solutions more quickly than unidirectional searches. Afterward, it merges the start tree and goal tree from the dual-tree search into a single tree to perform informed sampling, optimizing the current solution. Additionally, Al-Ansarry et al.~\cite{al2021hybrid} introduce an enhanced approach known as Hybrid RRT-A*, designed to address the limitations of the original RRT, particularly its slow convergence and high cost. By integrating the heuristic function of the A* algorithm with RRT, the method reduces tree expansion and directs the search towards the goal more efficiently, using fewer nodes and less time. Respectively, Kiani et al.~\cite{kiani2021adapted} integrate RRT with Grey Wolf Optimization (GWO), demonstrating the ability to efficiently find near-optimal paths in three-dimensional path planning. Pohan et al.~\cite{pohan2023efficient} propose a hybridization of RRT* and Ant Colony System (RRT-ACS)to generate optimal parking paths efficiently, outperforming other algorithms in common scenarios. Recently, Cao et al.~\cite{cao20223d} integrate the advantages of RRT-Connect~\cite{Kuffner2000rrt} for global path planning, artificial potential field (APF)~\cite{khatib1986real} for local path planning, and cubic B-spline~\cite{wang2011cubic} for curve smoothing to optimize the path of unmanned aerial vehicles (UAVs).

\section{Application and Limitation}
\label{sec:limitation}
\subsection{Application 1: Dynamic Environments}
Dynamic path planning involves determining a path from a starting position to a goal position in an environment that evolves over time. This may include changes caused by moving obstacles, such as individual pedestrians or crowds, alterations in the environment's structure, or other dynamic elements requiring real-time path updates~\cite{Bing2023TPAMI}. Various sampling-based algorithms have been developed to efficiently replan paths in such scenarios. One common approach is to use bidirectional search~\cite{zhang2022local, Zou2023dynamic}, which facilitates quicker updates. Another strategy focuses on continuously refining and adjusting the search process during execution~\cite{seif2015mobile, connell2017dynamic, Qi2021mod, zhao2023dynamic}. Furthermore, biased sampling has proven to be an effective method for improving planning efficiency~\cite{Adiyatov2017a, wei2018method, cai2019adaptive, yuan2020efficient}. Additionally, incorporating human awareness and crowd dynamics into the sampling-based planning process further enhances the planner's performance~\cite{chi2017risk,chi2018risk,cai2023sampling,cai2022human}.


\subsection{Application 2: Unknown Environments and Uncertainty Localization}

Robot motion planning in unknown environments is challenging, as it requires navigating without prior knowledge of the surroundings. In such cases, robots rely on real-time sensor data (e.g., LiDAR, cameras, sonar) to detect obstacles, track changes, and build an on-the-fly map. This process involves continuous exploration, obstacle avoidance, and dynamic path adjustments. Sampling-based algorithms play a crucial role in exploration tasks, enabling efficient and adaptive navigation in uncertain environments.

In unknown environments, one effective method involves using sampling-based algorithms with a biased direction~\cite{Tian2007Application,chang2008mobile} to optimize node generation and reduce computational load, allowing for more efficient real-time path planning. Other strategies include rolling planning combined with node screening~\cite{li2022improved} and model predictive control to further enhance path optimization~\cite{lindqvist2024tree}.

To address localization uncertainties, sampling-based algorithms integrated with the Extended Kalman Filter (EKF)\cite{pepy2006safe} and Simultaneous Localization and Mapping (SLAM)\cite{huang2008rrt} have been employed to improve navigation accuracy. Additionally, sampling-based methods that focus on generating robust trajectories~\cite{luders2014optimizing} and risk-bounded trajectories~\cite{summers2018distributionally} provide enhanced path-planning capabilities in uncertain environments. The Min-Max RRT*~\cite{englot2016sampling} further refines this process by minimizing the maximum state estimate uncertainty along a path, offering an alternative to traditional additive cost representations of uncertainty.


\subsection{Application 3: Nonholonomic constraints}
Nonholonomic constraints limit a robot's movement in certain directions due to its physical structure, such as wheeled robots that cannot move sideways. These constraints complicate motion planning, requiring algorithms that generate paths respecting the robot's kinematic properties. To handle these challenges, sampling-based planners are adapted using techniques like steering functions~\cite{banzhaf2017hybrid,peng2021towards,reyes2023visual}, and goal bias~\cite{gan2021research,dong2018faster}. Another approach combines RRT* with the generalized velocity obstacles (GVO) model to reduce trajectory uncertainty~\cite{chen2018rrt}. Similarly, the use of a distance function extends RRT* to handle nonholonomic constraints effectively~\cite{park2015feedback}. S-BRRT*~\cite{Zhang2021s} further optimizes planning by combining the strengths of bidirectional RRT* with pruning and smoothing strategies. It incorporates bidirectional trees to enhance search efficiency and employs Bezier curves for path smoothing, significantly improving path quality and exploration efficiency in both sparse and dense environments.

\subsection{Application 4: Kinematics}

Sampling-based motion planning with kinodynamic constraints considers both the robot's kinematics and dynamics, planning in the state space (including position and velocity). This approach requires generating time-parameterized, dynamically feasible trajectories, making the problem more complex.

A wide variety of sampling-based algorithms for kinodynamic planning exist. For instance, using a fixed-final-state-free-final-time controller~\cite{webb2012kinodynamic}, neural network predictions of cost functions to achieve the cost/metric between two given states considering the nonlinear constraints~\cite{Li2018neural}, deep reinforcement learning to learn an obstacle-avoiding
policy that maps a robot’s sensor observations to actions~\cite{chiang2019rl}, precomputed motion primitives~\cite{sakcak2019sampling}, lazy-steering techniques~\cite{yavari2019lazy} and using a partial-final-state-free (PFF) optimal controller in
kinodynamic RRT* to reduce the dimensionality of the
sampling space~\cite{zheng2021accelerating}.

\subsection{Application 5: Manifold constraints}
Motion planning under manifold constraints in robotics involves finding paths or trajectories that not only guide the robot from start to goal but also satisfy specific constraints defining a "manifold." Manifolds are higher-dimensional spaces representing the set of valid configurations that the robot can assume under its constraints, such as non-holonomic constraints, geometric constraints, kinematic limitations, or contact constraints.

Several sampling-based algorithms have been proposed to address these challenges, such as projection methods~\cite{berenson2009manipulation,suh2011tangent,suh2011tangent}, continuation techniques~\cite{jaillet2012path,jaillet2013efficient,jaillet2017path,kim2016tangent}, reparameterization-based~\cite{han2008convexly,mcmahon2016sampling} and offline methods that construct an approximation of the constraint manifold offline~\cite{csucan2012motion,burget2013whole}.



\subsection{Application 6: Multi-Robot}
Multi-robot motion planning focuses on coordinating the movements of multiple robots to complete individual or collective tasks while avoiding collisions. This involves not only determining paths for each robot within a shared environment but also managing their interactions to optimize overall efficiency. Such planning is crucial in applications like warehouse automation, search and rescue missions, autonomous bio-inspired robot coordination~\cite{Bing2023SR}, and robotic swarms.

Research in sampling-based algorithms has extensively tackled the challenges inherent in multi-robot motion planning, proposing various modifications and enhancements to traditional methods for improved performance. One approach~\cite{cain2021mk} integrates both tightly-coupled and loosely-coupled strategies within its framework, allowing it to adapt to diverse multi-robot task scenarios. Another method~\cite{hvvezda2019improved} utilizes a more efficient sampling strategy to enhance planning efficiency. Additionally, an optimization-based map exploration strategy~\cite{zhang2020rapidly} equips multiple robots to actively explore their environment.

To efficiently plan within composite spaces, sampling-based algorithms offer effective strategies, where constructing individual roadmaps for each robot and implicitly searching the tensor product of these structures is a common approach~\cite{adler2015efficient,solovey2016finding,shome2020drrt}.  For resolving inter-robot conflicts, sampling-based methods  incorporate techniques like Prioritized Planning (SI-CPP) and Conflict-Based Search (SI-CCBS)~\cite{sim2024safe} to better coordinate the paths of multiple robots.

\subsection{Limitation 1: Narrow passage}
In robot motion planning, the narrow passage problem refers to the challenge robots face when navigating through tight or confined spaces within complex environments.
Research on sampling-based algorithms has extensively explored the narrow passage problem. Various studies have proposed modifications and enhancements to the traditional sampling-based algorithms to improve their performance in these constrained environments. One approach focuses on optimizing the sampling process to enhance the accuracy and efficiency of path planning in narrow spaces~\cite{Zhou2023an, belaid2022narrow, chai2022rj}. Another strategy involves incorporating an improved bridge test along with a novel search method based on local guidance to facilitate smoother navigation~\cite{shu2019locally}. Additionally, Szkandera et al.\cite{szkandera2020narrow} propose a method inspired by the exit points for cavities in protein models, incorporating this concept into sampling-based algorithms. Respectively, Wang et al.~\cite{wang2018learning} model the tree selection process as a multi-armed bandit problem and using a reinforcement learning algorithm with an enhanced $\varepsilon_t$-greedy strategy to address the narrow passage problem.

\subsection{Limitation 2: Slow convergence and large memory required}
Sampling-based algorithms face several limitations and challenges, particularly related to low convergence rates and high memory requirements.  \textbf{Low Convergence Rates}: sampling-based algorithms can struggle with convergence speed, especially in high-dimensional spaces. The random sampling process, while effective for exploration, can lead to slow convergence towards optimal paths. This is a significant drawback when quick and efficient path planning is required. \textbf{High Memory Requirements}: sampling-based algorithms often require large amounts of memory to store the extensive tree structures generated during the planning process. This can become problematic in scenarios with limited computational resources or when dealing with very large state spaces.

To improve the convergence speed and memory efficiency of sampling-based algorithms, researchers have introduced various advanced methods. These include smart sampling techniques~\cite{Nasir2013RRT*-SMART, Islam2012rrt}, bidirectional search~\cite{Kuffner2000rrt}, lazy search strategies~\cite{hauser2015lazy}, and heuristic-based methods~\cite{Gammell2014informed, gammell2018informed}.

\subsection{Comprehensive Comparison of Sampling-based Algorithms (1998-2024)}
\begin{table*}[htp]
\centering
\renewcommand{\arraystretch}{1.4} 
\begin{adjustbox}{max width=\textwidth}
\begin{tabular}{>{\centering\arraybackslash}p{3cm} >{\centering\arraybackslash}p{1cm} >{\centering\arraybackslash}p{2.4cm} >{\centering\arraybackslash}p{2cm} >{\centering\arraybackslash}p{1.5cm} >{\centering\arraybackslash}p{1.2cm} >{\centering\arraybackslash}p{1.5cm} >{\centering\arraybackslash}p{1.5cm} >{\centering\arraybackslash}p{1.8cm} >{\centering\arraybackslash}p{1.2cm} >{\centering\arraybackslash}p{1.9cm} >{\centering\arraybackslash}p{1.8cm} >{\centering\arraybackslash}p{1.8cm}}
\hline
\hline
\textbf{Method} & \textbf{Year} & \textbf{Probabilistic Completeness} & \textbf{Bidirectional Approach} & \textbf{Heuristic Search} & \textbf{Batch Sampling} & \textbf{Asymptotic Optimal} & \textbf{Anytime Capability} & \textbf{Lazy Collision Check} & \textbf{Greedy Strategy} & \textbf{Learning-Based Planning} & \textbf{Static Path Planning} & \textbf{Dynamic Path Planning} \\
\hline
RRT~\cite{lavalle1998rapidly} & 1998 & \checkmark &  &  &  &  &  &  &  &  & \checkmark &  \\

RRT-connect~\cite{Kuffner2000rrt} & 2000 & \checkmark & \checkmark  & \checkmark  &  &   &  &  &  &  & \checkmark &  \\

Lazy PRM~\cite{Bohlin2000path} & 2000 & \checkmark &  & \checkmark &  &  &  & \checkmark &  &  & \checkmark &  \\

hRRT~\cite{urmson2003approaches} & 2003 & \checkmark &  & \checkmark &  &  &  &  &  &  & \checkmark &  \\

ADD-RRT~\cite{Jaillet2005adaptive} & 2005 & \checkmark &  & \checkmark &  &  &  &  &  &  & \checkmark &  \\

Anytime RRT~\cite{Ferguson2006anytime} & 2006 & \checkmark &  & \checkmark &  &  & \checkmark &  &  &  & \checkmark & \checkmark \\

T-RRT~\cite{Jaillet2008transition} & 2008 & \checkmark &  & \checkmark &  &  &  &  &  &  & \checkmark &  \\

TSRRT~\cite{um2010tangent} & 2010 & \checkmark & \checkmark & \checkmark &  &  &  & \checkmark &  &  & \checkmark &  \\

RRT*~\cite{Karaman2011sampling} & 2011 & \checkmark &  & \checkmark &  & \checkmark  & \checkmark &  &  &  & \checkmark &  \\

RRT*-SMART~\cite{Nasir2013RRT*-SMART} & 2013 & \checkmark &  & \checkmark &  & \checkmark & \checkmark &  &  &  & \checkmark &  \\

{RRT}$^\#$~\cite{arslan2013use} & 2013 & \checkmark &  & \checkmark &  & \checkmark & \checkmark &  &  &  & \checkmark & \checkmark \\

Informed RRT*~\cite{Gammell2014informed} & 2014 & \checkmark &  & \checkmark &  & \checkmark & \checkmark &  &  &  & \checkmark &  \\

FMT*~\cite{janson2015fastmarchingtreefast} & 2015 & \checkmark &  & \checkmark & \checkmark & \checkmark &  & \checkmark &  &  & \checkmark & \\

BIT*~\cite{gammell2015batch} & 2015 & \checkmark &  & \checkmark & \checkmark & \checkmark & \checkmark &  &  &  & \checkmark &  \\

Lazy-PRM*~\cite{hauser2015lazy} & 2015 & \checkmark &  & \checkmark &  & \checkmark & \checkmark & \checkmark &  &  & \checkmark & \checkmark \\

Lazy-RRG*~\cite{hauser2015lazy} & 2015 & \checkmark &  & \checkmark &  & \checkmark & \checkmark & \checkmark &  &  & \checkmark & \checkmark \\

RT-RRT*~\cite{naderi2015rt} & 2015 & \checkmark &  & \checkmark &  & \checkmark & \checkmark &  &  &  & \checkmark & \checkmark \\

B{I$^2$}RRT*~\cite{burget2016bi}  & 2016 & \checkmark & \checkmark & \checkmark &  & \checkmark & \checkmark &  & \checkmark &  & \checkmark & \\

RR{T$^X$}~\cite{Otte2016rrtx} & 2016 & \checkmark &  & \checkmark &  & \checkmark & \checkmark &  &  &  & \checkmark & \checkmark \\

FCRRT~\cite{kang2016fast} & 2016 & \checkmark &  & \checkmark &  & \checkmark &  & \checkmark &  &  & \checkmark & \checkmark \\

HARRT*~\cite{Yi2016homotopy} & 2016 & \checkmark & \checkmark & \checkmark &  & \checkmark & \checkmark &  &  &  & \checkmark & \checkmark \\

P-RRT*~\cite{Qureshi2016potential} & 2016 & \checkmark &  & \checkmark &  & \checkmark & \checkmark &  &  &  & \checkmark &  \\

ALCC~\cite{kim2018adaptive} & 2018 & \checkmark &  & \checkmark &  & \checkmark & \checkmark & \checkmark &  &  & \checkmark & \checkmark \\

NoD-RRT~\cite{Li2018neural} & 2018 & \checkmark & & \checkmark &  & \checkmark & \checkmark &  &  & \checkmark & \checkmark &  \\

PB-RRT*~\cite{tahir2018potentially} & 2018 & \checkmark & \checkmark & \checkmark &  & \checkmark & \checkmark &  &  &  & \checkmark &  \\

PIB-RRT*~\cite{tahir2018potentially} & 2018 & \checkmark & \checkmark & \checkmark &  & \checkmark & \checkmark &  &  &  & \checkmark &  \\

Multi-Tree T-RRT*~\cite{wong2018optimal} & 2018 & \checkmark &  & \checkmark &  & \checkmark &  &  &  &  & \checkmark & \checkmark \\

Lazy Steering RRT*~\cite{yavari2019lazy} & 2019 & \checkmark &  & \checkmark &  & \checkmark & \checkmark & \checkmark &  & \checkmark & \checkmark &  \\

RL-RRT~\cite{chiang2019rl} & 2019 & \checkmark &  & \checkmark &  &  &  &  &  & \checkmark & \checkmark &  \\

Quick-RRT*~\cite{jeong2019quick} & 2019 & \checkmark &  & \checkmark &  & \checkmark & \checkmark &  &  &  & \checkmark &  \\

PQ-RRT*~\cite{li2020pq} & 2020 & \checkmark &  & \checkmark &  & \checkmark & \checkmark &  &  &  & \checkmark &  \\

AIT*~\cite{strub2020adaptively} & 2020 & \checkmark & \checkmark & \checkmark & \checkmark  & \checkmark & \checkmark & \checkmark &  &  & \checkmark & \\

ABIT*~\cite{strub2020advanced} & 2020 & \checkmark &  & \checkmark & \checkmark & \checkmark & \checkmark &  &  &  & \checkmark &  \\

dRRT*~\cite{shome2020drrt} & 2020 & \checkmark &  & \checkmark &  & \checkmark & \checkmark &  &  &  & \checkmark & \checkmark \\

Neural RRT*~\cite{Wang2020neural} & 2020 & \checkmark &  & \checkmark &  & \checkmark & \checkmark &  &  & \checkmark & \checkmark &  \\

GBIT*~\cite{chen2021greedy} & 2021 & \checkmark &  & \checkmark & \checkmark & \checkmark & \checkmark & & \checkmark &  & \checkmark &  \\

{MP-RRT}$^\#$~\cite{Primatesta2021amodel} & 2021 & \checkmark &  & \checkmark &  &  & \checkmark &  &  &  & \checkmark & \checkmark \\

AM-RRT*~\cite{Armstrong2021amrrt} & 2021 & \checkmark &  & \checkmark &  & \checkmark & \checkmark &  &  &  & \checkmark & \checkmark \\

MOD-RRT*~\cite{Qi2021mod} & 2021 & \checkmark &  & \checkmark &  & \checkmark & \checkmark &  &  &  & \checkmark & \checkmark \\

F-RRT*~\cite{Liao2021f} & 2021 & \checkmark &  & \checkmark &  &  & \checkmark &  &  &  & \checkmark & \checkmark \\

EIT*~\cite{strub2022adaptively} & 2022 & \checkmark &\checkmark  & \checkmark & \checkmark & \checkmark & \checkmark & \checkmark &  &  & \checkmark & \\

EIRM*~\cite{hartmann2022effort} & 2022 & \checkmark & \checkmark & \checkmark & \checkmark & \checkmark & \checkmark & \checkmark &  &  & \checkmark & \\

RJ-RRT~\cite{chai2022rj} & 2022 & \checkmark &  & \checkmark &  &  &  &  & \checkmark &  & \checkmark & \\

BiAIT*~\cite{Li2023biait} & 2023 & \checkmark & \checkmark & \checkmark & \checkmark & \checkmark & \checkmark & \checkmark &  &  & \checkmark &  \\

EP-RRT*~\cite{ding2023improved} & 2023 & \checkmark & \checkmark & \checkmark &  & \checkmark & \checkmark &  & \checkmark &  & \checkmark &  \\

FA-RRT*N~\cite{Khattab2023intelligent} & 2023 & \checkmark &  & \checkmark &  & \checkmark & \checkmark &  &  &  & \checkmark &  \\

GMM-RRT*~\cite{Lv2023gmm} & 2023 & \checkmark &  & \checkmark &  & \checkmark  & \checkmark &  &  &  & \checkmark & \\

PROTAMP-RRT~\cite{saccuti2023protamp} & 2023 & \checkmark &  & \checkmark &  &  &  &  &  &  & \checkmark & \\

Neural Informed RRT*~\cite{huang2024neural} & 2024 & \checkmark &  & \checkmark &  & \checkmark & \checkmark &  &  & \checkmark & \checkmark & \checkmark \\

iDb-RRT~\cite{ortizharo2024idbrrt} & 2024 & \checkmark & \checkmark & \checkmark &  &  &  &  &  &  & \checkmark &  \\

SI-RRT*~\cite{sim2024safe} & 2024 & \checkmark &  & \checkmark &  & \checkmark & \checkmark &  &  &  & \checkmark & \checkmark \\
\hline
\end{tabular}
\end{adjustbox}
\caption{Key Characteristics and Algorithmic Features of Sampling-based Planners}
\end{table*}

This table provides a comprehensive comparison of various typical sampling-based algorithms proposed between 1998 and 2024, showcasing the evolution of path planning techniques over the years. The table lists key attributes for each algorithm, including probabilistic completeness, bidirectional approach, heuristic search, batch sampling, asymptotic optimality, anytime capability, lazy collision check, greedy strategy, learning-based planning, static path planning, and dynamic path planning. These attributes highlight each method's technical advancements and specific features, providing insight into how they address different path planning challenges.
\section{Experiment}
\label{sec:experiment}

In this survey, we utilize the Planner-Arena benchmark database~\cite{moll2015benchmarking}, the Planner Developer Tools (PDT)~\cite{gammell2022planner}, and Open Robotics Automation Virtual Environment (OpenRAVE)~\cite{Diankov2010} to benchmark proposed motion planner behaviors. 

\begin{figure}[htp]
    \centering
    \begin{tikzpicture}
    \node[inner sep=0pt] (russell) at (-4.0,0.0)
    {\includegraphics[width=0.248\textwidth]{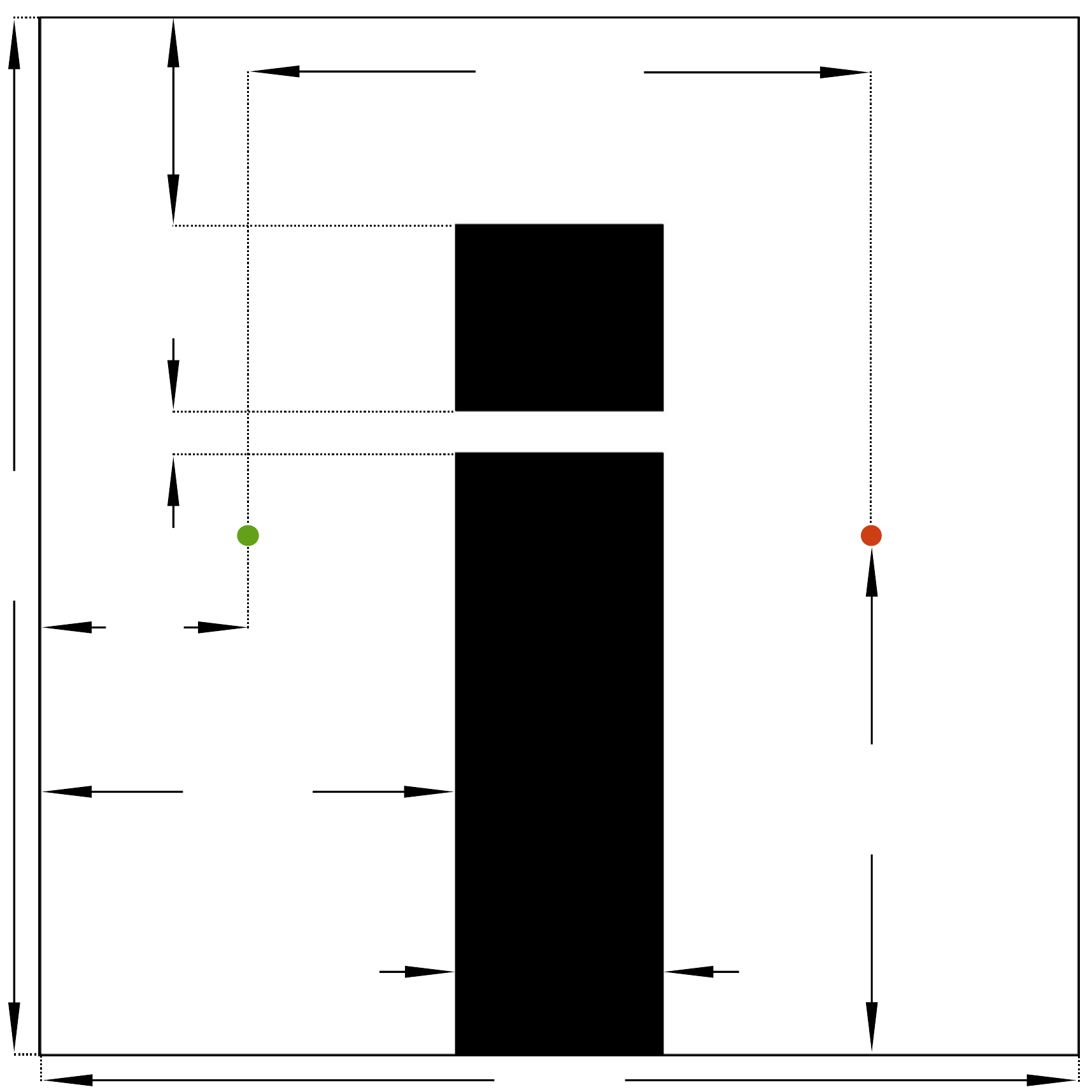}};
    \node[inner sep=0pt] (russell) at (0.5,0.0)
    {\includegraphics[width=0.248\textwidth]{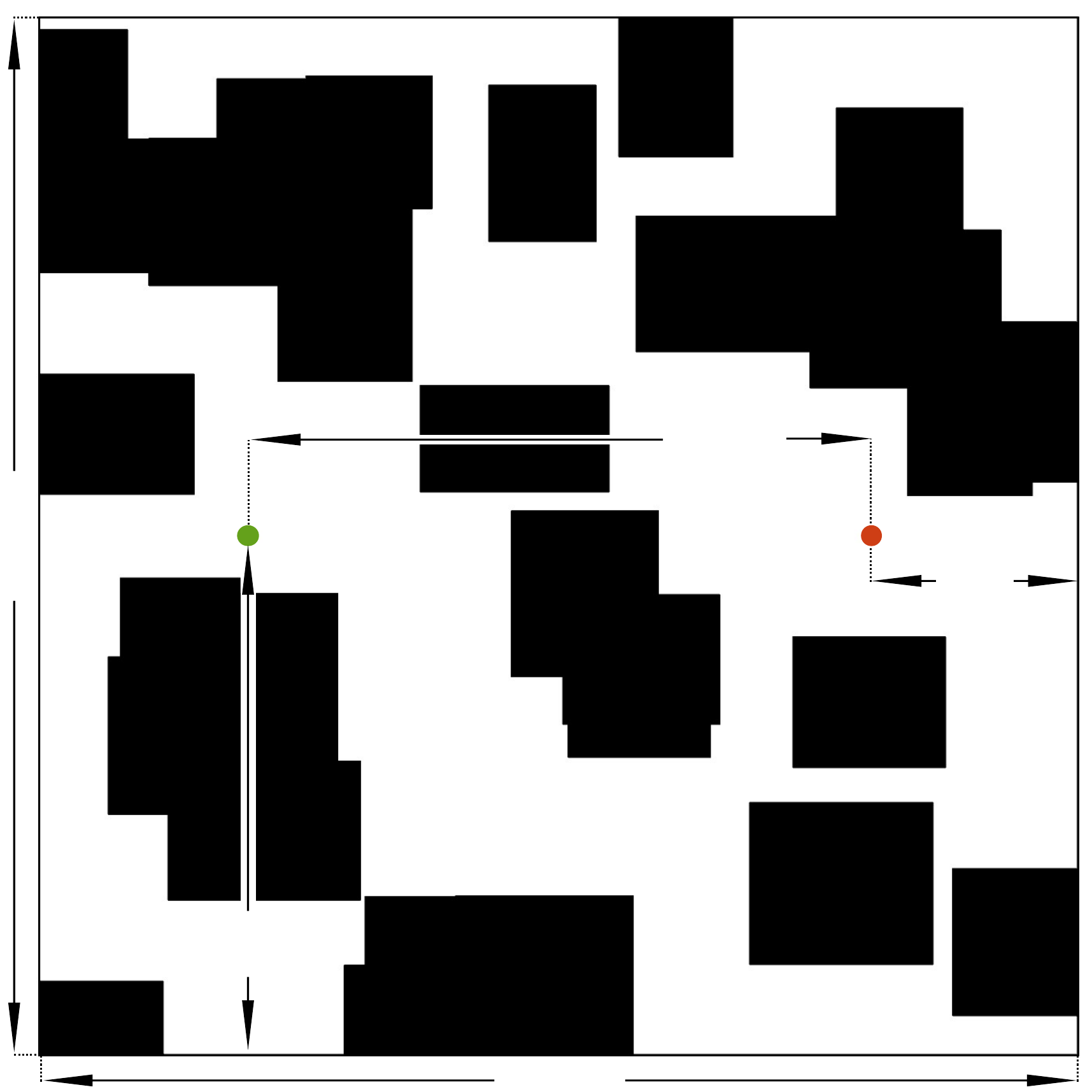}};
    \scriptsize
    
    \node at (-3.97,1.97) {0.6};
    \node at (-5.66,-0.32) {0.2};
    \node [rotate=90] at (-5.7,0.72) {0.04};    
    \node [rotate=90] at (-5.7,1.72) {0.2};
    \node [rotate=90] at (-2.65,-1.04) {0.5};
    \node at (-5.24,-1.01) {0.4};
    \node at (-4.92,-1.75) {0.2};
    \node [rotate=90] at (-6.2,0.05) {1.0};
    \node at (-4.0,-2.22) {1.0};
    \node at (-4.8,0.1) {\color{teal} Start};
    \node at (-3.06,0.1) {\color{purple} Goal};
    
    \node at (1.22,0.43) {0.6};
    \node at (2.28,-0.14) {0.2};
    \node [rotate=90] at (-0.69,-1.62) {0.5};
    \node [rotate=90] at (-1.7,0.05) {1.0};
    \node at (0.5,-2.22) {1.0};
    \node at (-0.3,0.1) {\color{teal} Start};
    \node at (1.44,0.1) {\color{purple} Goal};

    \node at (-4.0,-2.6) {\small (a) Wall Gap (WG)};
    \node at (0.5,-2.6) {\small (b) Random Rectangles (RR)};
    \end{tikzpicture}
    \caption{(a) A two-dimensional illustration of the wall gap experiment. The start and goal states are represented by a green dot and red dot, respectively. Each state space dimension was bounded to the interval [0, 1]. (b) A two-dimensional illustration of the Random Rectangles experiment. The start and goal states are represented by a green dot and red dot, respectively. Each state space dimension was bounded to the interval [0, 1]. }
    \label{fig: testEnv}
    \vspace{0.5em} 
\end{figure}
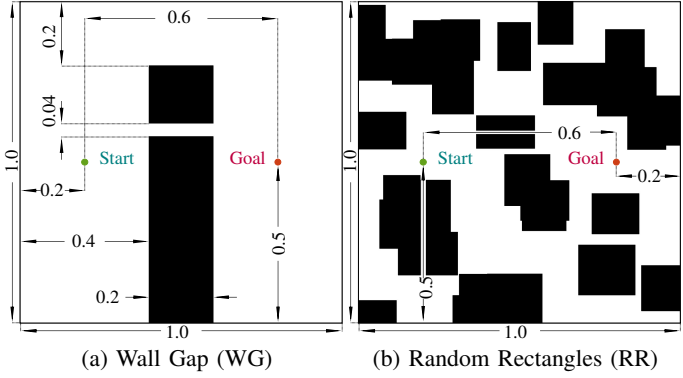

\begin{figure*}[htp]
    \centering
    \begin{tikzpicture}
    \node[anchor=center] at (0,0) 
    {\includegraphics[width=0.95\textwidth]{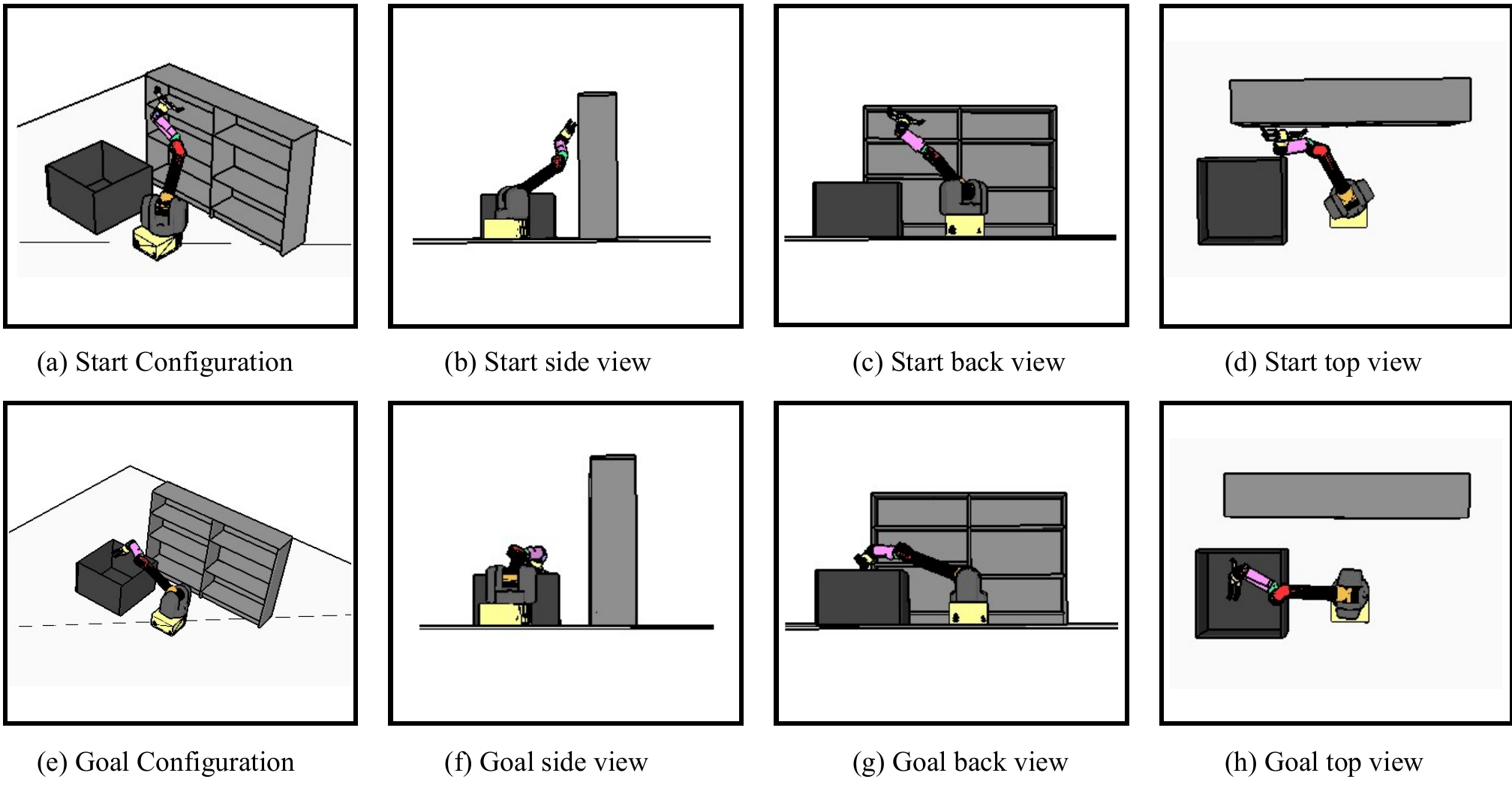}};
    \end{tikzpicture}
    \caption{Illustrations of the single-arm manipulator problem. The top row shows the start configuration of the arm in position to pick up an object on the top shelf (a–d). The bottom row shows the goal configuration of the arms in position to place an object in the
    box (e–h). }
    \label{fig:oneArm}
    \vspace{0.5em}
\end{figure*}

\begin{figure*}[htp]
    \centering
    \begin{tikzpicture}
    \node[anchor=center] at (0,0) 
    {\includegraphics[width=0.95\textwidth]{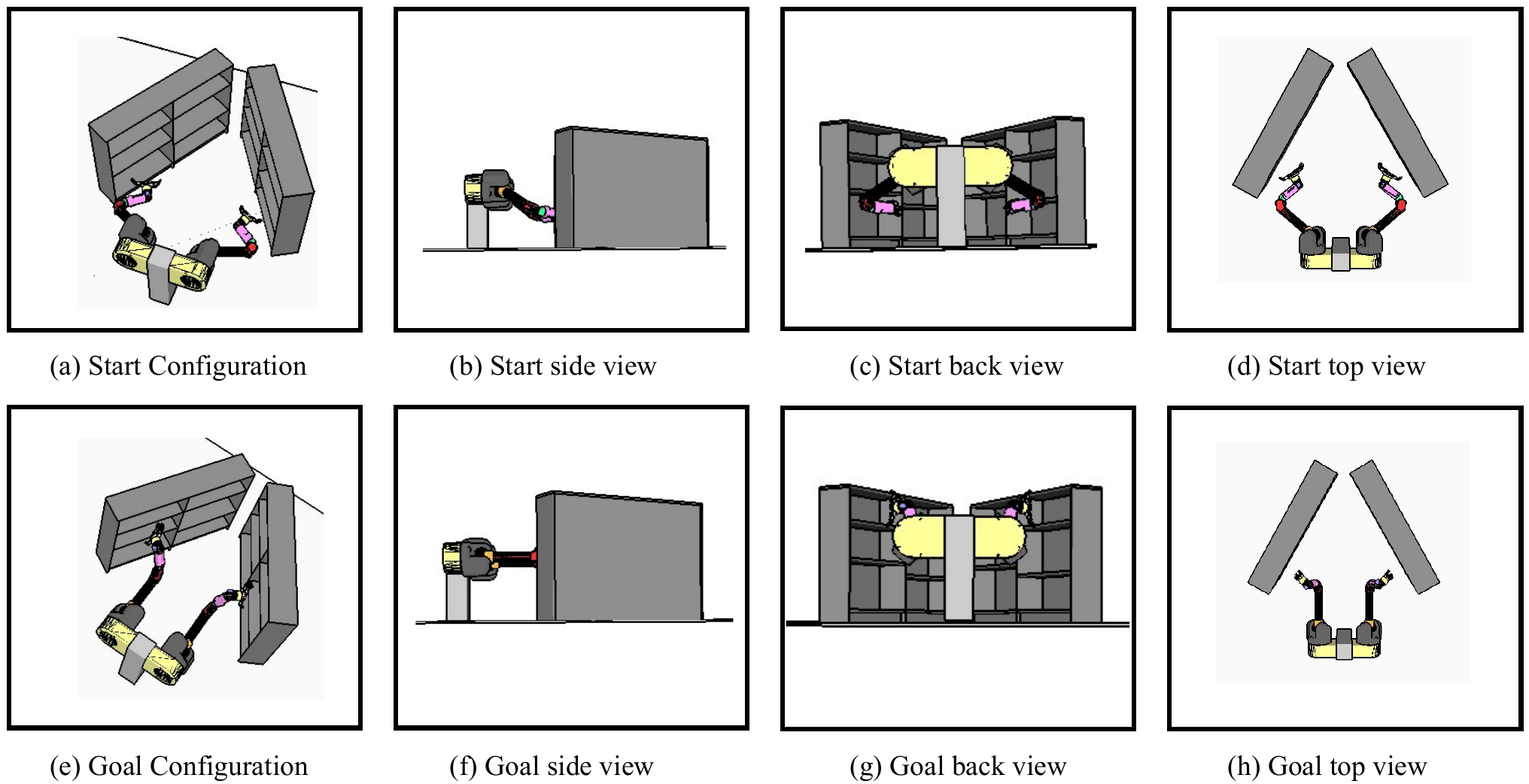}};
    \end{tikzpicture}
    \caption{Illustrations of the dual-arm manipulator problem. The top row shows the start configuration of the arms in position to pick up an object on the bottom shelf (a–d). The bottom row shows the goal configuration of the arms in position to place an object on the
    top shelf (e–h). }
    \label{fig:dualArm}
    \vspace{0.5em}
\end{figure*}

Ten popular algorithms, including various versions of RRT-Connect, RRT*, RRT$^\#$, Informed RRT*, LazyPRM*, BIT*, ABIT*, AIT*, EIRM*, and EIT* from the Open Motion Planning Library (OMPL)\cite{sucan2012open}, were tested in both simulated random scenarios (Fig.\ref{fig: testEnv}) and manipulation tasks (Fig.~\ref{fig:oneArm}, Fig.~\ref{fig:dualArm}).
The evaluations are implemented on a desktop with an Intel i5-12600k processor and 16GB memory, running Ubuntu 20.04. These comparisons were carried out in simulated environments ranging from $\mathbb{R}^4$ to $\mathbb{R}^{16}$, and for Manipulation tasks ranging from $\mathbb{R}^7$ and $\mathbb{R}^{14}$. The primary objective for the planners was to minimize path length (cost). The RGG constant $\eta$ was uniformly set to 1.001, and the rewire factor was set to 1.2 for all planners.

For RRT-based algorithms, a 5\% goal bias was used, with maximum edge lengths of 0.5, 1.1, 1.25, 2.4 and 3.0 in $\mathbb{R}^4$, $\mathbb{R}^7$, $\mathbb{R}^8$, $\mathbb{R}^{14}$,$\mathbb{R}^{16}$. All batch-sorted planners sampled 100 states per batch, and informed planners defined the informed set $X_{\hat{f}}$ using the current best costs.

\subsection{Simulation Experimental Tasks}

\begin{figure*}[htp]
    \centering
    \begin{tikzpicture}
    \node[inner sep=0pt] (russell) at (4.1,9)
    {\includegraphics[width=0.4\textwidth]{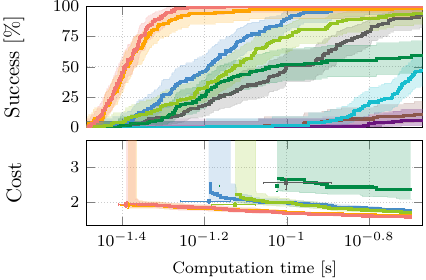}};
    \node[inner sep=0pt] (russell) at (4.1,3.5)
    {\includegraphics[width=0.4\textwidth]{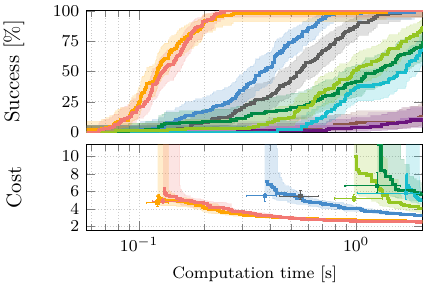}};
    \node[inner sep=0pt] (russell) at (4.1,-2)
    {\includegraphics[width=0.4\textwidth]{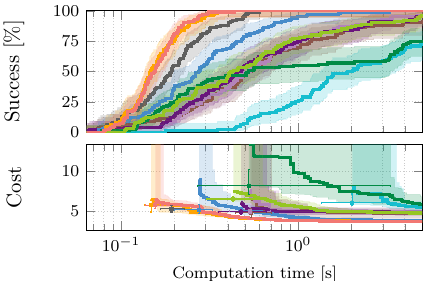}};

    \node[inner sep=0pt] (russell) at (-4.9,9)
    {\includegraphics[width=0.4\textwidth]{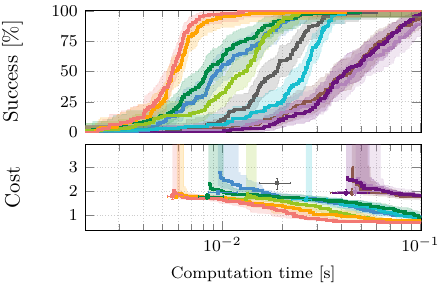}};
    \node[inner sep=0pt] (russell) at (-4.9,3.5)
    {\includegraphics[width=0.4\textwidth]{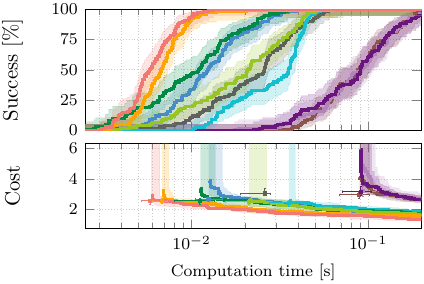}};  
    \node[inner sep=0pt] (russell) at (-4.9,-2){\includegraphics[width=0.4\textwidth]{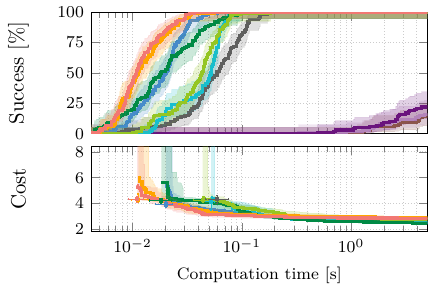}};

    \node[inner sep=0pt] (russell) at (0.0,-5.5){\includegraphics[width=0.92\textwidth]{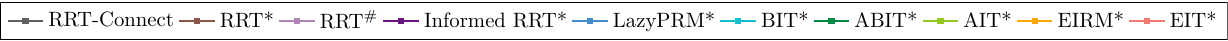}};

    \node at (-4.5,6.3) {\footnotesize (a) WG in $\mathbb{R}^4$ - MaxTime: 0.10s};
    \node at (-4.5,0.8) {\footnotesize(c) WG in $\mathbb{R}^8$ - MaxTime: 0.20s};
    \node at (-4.5,-4.8) {\footnotesize(e) WG in $\mathbb{R}^{16}$ - MaxTime: 5.00s};

    \node at (4.5,6.3) {\footnotesize (b) RR in $\mathbb{R}^4$ - MaxTime: 0.20s};
    \node at (4.5,0.8) {\footnotesize(d) RR in $\mathbb{R}^8$ - MaxTime: 2.00s};
    \node at (4.5,-4.8) {\footnotesize(f) RR in $\mathbb{R}^{16}$ - MaxTime: 5.00s};

    \end{tikzpicture}
    \caption{Detailed experimental results from various motion planners in the Open Motion Planning Library are presented above. Figures (a), (c) and (e) depict test benchmark wall gap outcomes in $\mathbb{R}^4$, $\mathbb{R}^8$ and $\mathbb{R}^{16}$, respectively. Panel (b) showcases ten random rectangle experiments in $\mathbb{R}^4$, while panels (d) and (f) demonstrate in $\mathbb{R}^8$ and $\mathbb{R}^{16}$. In the cost plots, subfigures illustrate both the solution cost and time, while lines indicate the progression of cost for a planner that is almost certainly asymptotically optimal (unsuccessful runs are assigned an infinite cost). The error bars depict nonparametric 99\% confidence intervals for both solution cost and time.}
    \label{fig: simulation-result}
    \vspace{0.5em}
\end{figure*}
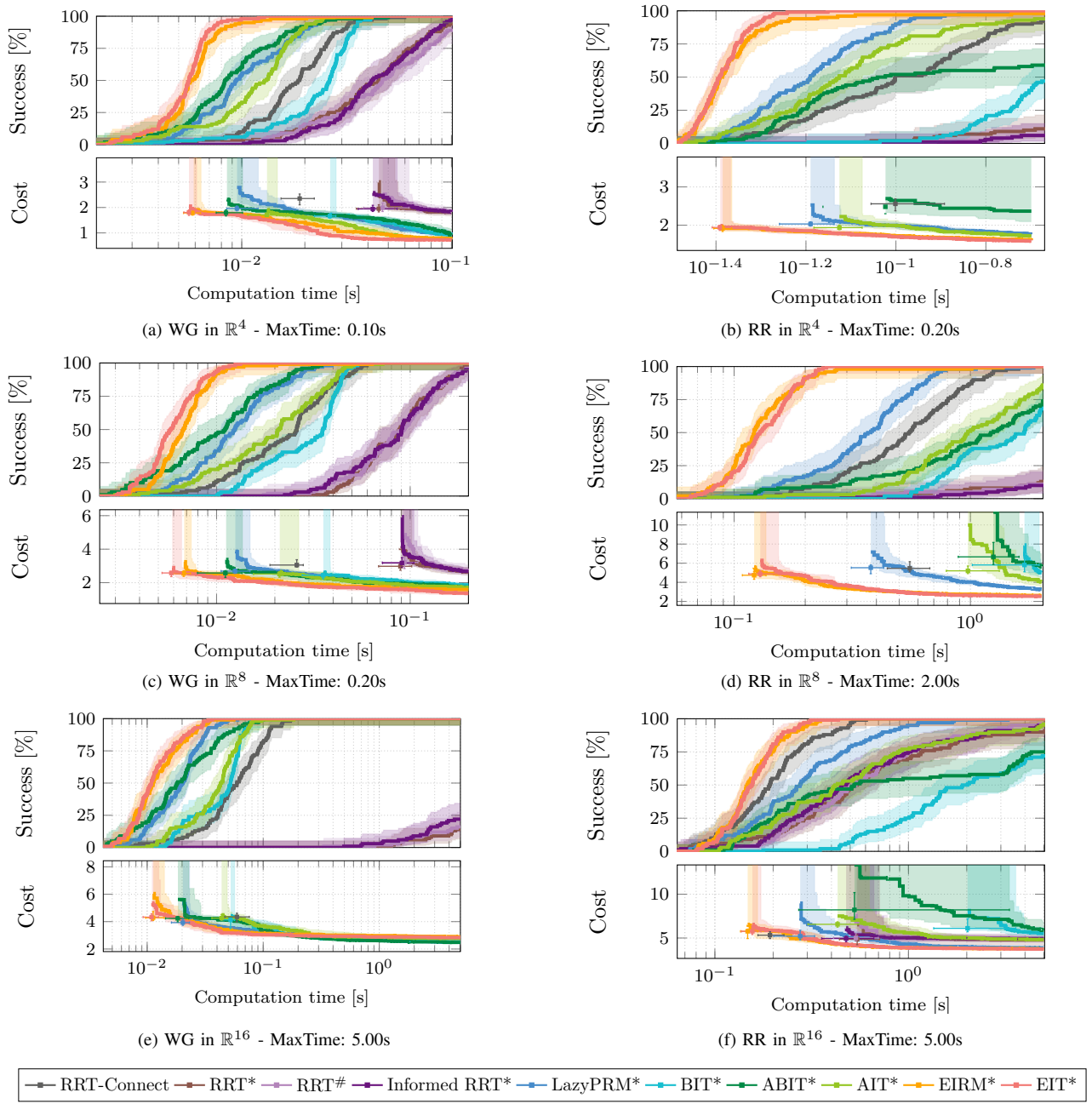

\begin{table*}[t!]
\caption{Benchmarks evaluation comparison for simulated random scenarios. \( t^\textit{med}_\textit{init} \): median initial solution time(in seconds); \( c^\textit{med}_\textit{init} \): median initial solution cost; \( c^\textit{med}_\textit{final} \): median solution cost over time(in seconds).}
\centering
\resizebox{0.92\textwidth}{!}{
\begin{tabular}{p{1.7cm}||c|c|c||c|c|c||c|c|c||c|c|c||c|c|c|c}
 \hline
 & \multicolumn{3}{c||}{\textbf{RRT-Connect}} & \multicolumn{3}{c||}{\textbf{RRT*}} & \multicolumn{3}{c||}{\textbf{RRT\#}} & \multicolumn{3}{c||}{\textbf{Informed RRT*}} & \multicolumn{3}{c|}{\textbf{LazyPRM*}} & \multirow{2}{*}
 
 \\
 
& \( t^\textit{med}_\textit{init} \) & \( c^\textit{med}_\textit{init} \) & \( c^\textit{med}_\textit{final} \) & \( t^\textit{med}_\textit{init} \) & \( c^\textit{med}_\textit{init} \) & \( c^\textit{med}_\textit{final} \) & \( t^\textit{med}_\textit{init} \) & \( c^\textit{med}_\textit{init} \) & \( c^\textit{med}_\textit{final} \) & \( t^\textit{med}_\textit{init} \) & \( c^\textit{med}_\textit{init} \) & \( c^\textit{med}_\textit{final} \) & \( t^\textit{med}_\textit{init} \) & \( c^\textit{med}_\textit{init} \) & \( c^\textit{med}_\textit{final} \) & \\
 \hline
    \( \text{WG}-\mathbb{R}^4 \) & 0.0189 & 2.3523  & 2.3523  & 0.0450  & 1.9527  & 1.8084  & 0.0469 & 1.9886 & 1.8418  & 0.0422  & 1.9418 & 1.8294 & 0.0095 & 1.9608 & 0.9560   \\
    \( \text{WG}-\mathbb{R}^8 \) & 0.0260  & 3.0528 & 3.0528 & 0.0891  & 2.9789 & 2.6418  & 0.0934  & 3.2095  & 2.7300  & 0.0907 & 3.1726  & 2.6795 & 0.0125 & 2.6882 & 1.6260  \\
    \( \text{WG}-\mathbb{R}^{16} \) & 0.0542 & 4.2013  & 4.2013 & 1.1494 &5.5114 &5.4170  & 1.3328 & 6.3678  & 6.0390  & 1.4954 & 5.3444 & 5.2526  & 0.0202 & 3.8372 & 2.8594  \\
    \( \text{RR}-\mathbb{R}^4 \) & 0.0997  & 2.5585 & 2.5585  & 0.2462  & 3.266  & 3.143  & 0.2480  & 3.077  & 3.356  & 0.2568 & 3.281  & 3.281 & 0.0646 & 2.0323 & 1.7571  \\
    \( \text{RR}-\mathbb{R}^8 \) & 0.5539  & 5.4617 & 5.4617  & 2.0761  & 7.2886  & 7.2435 & 2.1085  & 7.0050  & 7.0042  &  2.2842 & 7.6778  & 7.2683 & 0.3788 & 5.5239 & 3.2652  \\
    \( \text{RR}-\mathbb{R}^{16} \) & 0.1926  & 5.3251 & 5.3251 & 0.5433  & 7.8501 & 7.7856  & 0.5349  & 7.9862  & 7.0149  & 0.4763  & 7.9458  & 7.9198 & 0.2759 & 5.2091 & 3.8871  \\
 \hline
  \hline
& \multicolumn{3}{c||}{\textbf{BIT*}} & \multicolumn{3}{c||}{\textbf{ABIT*}} & \multicolumn{3}{c||}{\textbf{AIT*}} & \multicolumn{3}{c||}{\textbf{EIRM* }} & \multicolumn{3}{c|}{\textbf{EIT*}} & \multirow{2}{*}

\\

 & \( t^\textit{med}_\textit{init} \) & \( c^\textit{med}_\textit{init} \) & \( c^\textit{med}_\textit{final} \) & \( t^\textit{med}_\textit{init} \) & \( c^\textit{med}_\textit{init} \) & \( c^\textit{med}_\textit{final} \) & \( t^\textit{med}_\textit{init} \) & \( c^\textit{med}_\textit{init} \) & \( c^\textit{med}_\textit{final} \) & \( t^\textit{med}_\textit{init} \) & \( c^\textit{med}_\textit{init} \) & \( c^\textit{med}_\textit{final} \) & \( t^\textit{med}_\textit{init} \) & \( c^\textit{med}_\textit{init} \) & \( c^\textit{med}_\textit{final} \) & \\
\hline
    \( \text{WG}-\mathbb{R}^4 \) & 0.0263 & 1.6643 & 0.9223 & 0.0084 & 1.7950  & 0.9151 & 0.0132 & 1.7087 & 0.7539  & \textcolor{red}{0.0058}  & 1.7849 & 0.7812  & \textcolor{red}{0.0056} & 1.8059 & 0.7235  \\
    \( \text{WG}-\mathbb{R}^8 \) & 0.0355 & 2.4434 & 1.8654  & 0.0111 & 2.5784 & 1.7255 & 0.0211 & 2.5161  & 1.7029  & \textcolor{red}{0.0068} & 2.5528 & 1.5862 & \textcolor{red}{0.0058} & 2.5802 & 1.3561  \\
    \( \text{WG}-\mathbb{R}^{16} \) & 0.0533 & 4.0784 & 2.8080 & 0.0174 & 4.2886 & 2.8300  & 0.0369 & 4.2280  & 3.1773  & \textcolor{red}{0.0104} & 4.6266 & 3.0372  & \textcolor{red}{0.0102} & 4.2056 & 2.9879 \\
    \( \text{RR}-\mathbb{R}^4 \) & 0.1239 & 2.2457 & 2.0457 & 0.0947  & 2.4809 & 2.3671 & 0.0749 & 1.9368  & 1.7195  & \textcolor{red}{0.0412} & 1.9173  & 1.6124 & \textcolor{red}{0.0406} & 1.9417 & 1.5854  \\
    \( \text{RR}-\mathbb{R}^8 \) & 1.6933 & 5.8140 & 5.0191  & 1.2512  & 6.6619  & 5.6050  &0.9761  & 5.2000  & 4.0319  & \textcolor{red}{0.1215}  & 4.7398 & 2.6180 & \textcolor{red}{0.1290} & 4.8876 & 2.5272   \\
    \( \text{RR}-\mathbb{R}^{16} \) & 2.0198  & 6.1065 & 5.5245  & 0.5259  & 8.2124  & 5.8792  & 0.4301  & 6.5640  & 4.8204  & \textcolor{red}{0.1477} & 5.7654  & 3.7562 & \textcolor{red}{0.1561} & 5.7635 & 3.7288  \\
 \hline
\end{tabular}}
\label{tab:benchmark-simulation}
\end{table*}


Fig.~\ref{fig: simulation-result} shows the performance of all algorithms in the wall gap test benchmarks and ten random rectangle experiments conducted in $\mathbb{R}^4$, $\mathbb{R}^8$, and $\mathbb{R}^{16}$. EIT* and EIRM* perform efficiently in terms of both success rate and path cost across various scenarios. RRT*, {RRT}$^\#$, and Informed RRT* do not perform as well in terms of both success rate and path cost across various scenarios. RRT* and {RRT}$^\#$ Informed RRT* tend to struggle with convergence speed, often requiring more time to find feasible solutions. 
As shown in Table~\ref{tab:benchmark-simulation}, various path planning algorithms are compared across multiple benchmark scenarios. Among benchmarked planners, EIT* stands out as the state-of-the-art (SOTA) algorithm, achieves the lowest initial solution time of 0.0056s in the \( \text{WG}-\mathbb{R}^4 \) scenario. This impressive performance highlights EIT*'s efficiency in quickly finding feasible paths, outperforming other algorithms such as RRT-Connect, RRT*, and Informed RRT*, which exhibit longer initial solution times.

In the \( \text{RR}-\mathbb{R}^4 \) scenario, \textbf{BIT*} and \textbf{ABIT*} achieve shorter initial times (0.1239s and 0.0947s, respectively), compared to the other methods, with BIT* also yielding a lower final solution cost (\( c^\textit{med}_\textit{final} = 2.0457 \)). Similarly, in \( \text{RR}-\mathbb{R}^{16} \), ABIT* outperforms other planners in terms of initial time (0.5259s), although EIT* demonstrates a lower final cost. In higher-dimensional scenarios, EIT* continues to demonstrate its competitive edge, with notable efficiency observed in both \( \text{WG}-\mathbb{R}^8 \) and \( \text{WG}-\mathbb{R}^{16} \). The ability of EIT* to maintain low initial solution times while achieving satisfactory final solution costs reinforces its status as a leading approach in path planning.

Overall, the table underscores the adaptability of sampling-based planners in high-dimensional spaces, achieving low initial solution times and final costs across various scenarios. ABIT* and BIT* excel in structured environments such as \( \text{RR}-\mathbb{R}^4 \) and \( \text{RR}-\mathbb{R}^{16} \), effectively balancing initial time and solution cost. EIT* and EIRM* exhibit similarly low values for median initial solution time, median initial solution cost, and median solution cost over time. In contrast, RRT*, {RRT}$^\#$, and Informed RRT* demonstrate higher median initial solution times, initial solution costs, and solution costs over time.


\subsection{Simulated Manipulation Tasks}

\begin{figure*}[htp]
    \centering
    \begin{tikzpicture}
    \node[inner sep=0pt] (russell) at (-4.9,9)
    {\includegraphics[width=0.4\textwidth]{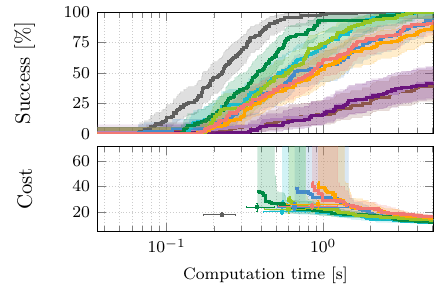}};

    \node[inner sep=0pt] (russell) at (4.1,9)
    {\includegraphics[width=0.45\textwidth]{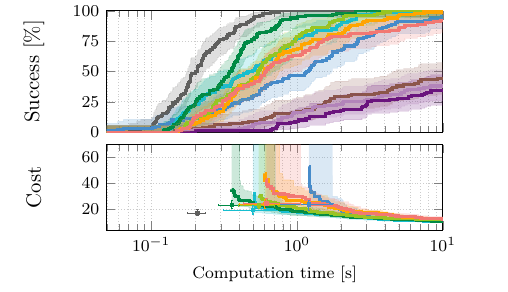}};

    \node[inner sep=0pt] (russell) at (0.0, 5){\includegraphics[width=0.92\textwidth]{figure/benchmark/rrt_10.pdf}};

    \node at (-4.5,6.3) {\footnotesize (a) Single-Arm (SA) in $\mathbb{R}^7$ - MaxTime: 5.0s};
    \node at (4.5,6.4) {\footnotesize (b) Dual-Arm (DA) in $\mathbb{R}^{14}$ - MaxTime: 10.0s};

    \end{tikzpicture}
    \caption{Detailed experimental results of the planners' performances on the single-arm and dual-arm manipulator are presented above. Figure (a) shows the result of the single-arm manipulator, Figure (b) shows the result of the dual-arm manipulator. In the cost plots, subfigures illustrate both the solution cost and time, while lines indicate the progression of cost for a planner that is almost certainly asymptotically optimal (unsuccessful runs are assigned an infinite cost). The error bars depict nonparametric 99\% confidence intervals for both solution cost and time. }
    \label{fig:result-openrave}
    \vspace{0.5em}
\end{figure*}

\begin{table*}[t!]
\caption{Benchmarks evaluation comparison for Manipulation tasks. SA: single-arm; DA: dual-arm; \( t^\textit{med}_\textit{init} \): median initial solution time(in seconds); \( c^\textit{med}_\textit{init} \): median initial solution cost; \( c^\textit{med}_\textit{final} \): median solution cost over time(in seconds).}
\centering
\resizebox{0.92\textwidth}{!}{
\begin{tabular}{p{1.7cm}||c|c|c||c|c|c||c|c|c||c|c|c||c|c|c|c}
 \hline
 & \multicolumn{3}{c||}{\textbf{RRT-Connect}} & \multicolumn{3}{c||}{\textbf{RRT*}} & \multicolumn{3}{c||}{\textbf{RRT\#}} & \multicolumn{3}{c||}{\textbf{Informed RRT*}} & \multicolumn{3}{c|}{\textbf{LazyPRM*}} & \multirow{2}{*}
 
 \\
& \( t^\textit{med}_\textit{init} \) & \( c^\textit{med}_\textit{init} \) & \( c^\textit{med}_\textit{final} \) & \( t^\textit{med}_\textit{init} \) & \( c^\textit{med}_\textit{init} \) & \( c^\textit{med}_\textit{final} \) & \( t^\textit{med}_\textit{init} \) & \( c^\textit{med}_\textit{init} \) & \( c^\textit{med}_\textit{final} \) & \( t^\textit{med}_\textit{init} \) & \( c^\textit{med}_\textit{init} \) & \( c^\textit{med}_\textit{final} \) & \( t^\textit{med}_\textit{init} \) & \( c^\textit{med}_\textit{init} \) & \( c^\textit{med}_\textit{final} \) & \\
 \hline
    \( \text{SA}-\mathbb{R}^7 \) & \textcolor{red}{0.2260} & 18.0909   & 18.0909  & 0.81   & 20.9410   & 20.9410   & 0.8561  & 20.7685  & 20.7685   & 0.818   & 24.4390  & 24.4390  & 0.6554  & 24.0031  & 13.2513    \\
    \( \text{DA}-\mathbb{R}^{14} \) & \textcolor{red}{0.2077}  & 16.7844  & 16.7844 & 1.440  & 16.8691 & 16.8691   & 1.502  & 24.7865   & 24.7865  & 1.514  & 24.7505   & 24.7505  & 1.2077  & 23.6651 & \textcolor{red}{11.9351}  \\
   
 \hline
  \hline
& \multicolumn{3}{c||}{\textbf{BIT*}} & \multicolumn{3}{c||}{\textbf{ABIT*}} & \multicolumn{3}{c||}{\textbf{AIT*}} & \multicolumn{3}{c||}{\textbf{EIRM* }} & \multicolumn{3}{c|}{\textbf{EIT*}} & \multirow{2}{*}

\\
 & \( t^\textit{med}_\textit{init} \) & \( c^\textit{med}_\textit{init} \) & \( c^\textit{med}_\textit{final} \) & \( t^\textit{med}_\textit{init} \) & \( c^\textit{med}_\textit{init} \) & \( c^\textit{med}_\textit{final} \) & \( t^\textit{med}_\textit{init} \) & \( c^\textit{med}_\textit{init} \) & \( c^\textit{med}_\textit{final} \) & \( t^\textit{med}_\textit{init} \) & \( c^\textit{med}_\textit{init} \) & \( c^\textit{med}_\textit{final} \) & \( t^\textit{med}_\textit{init} \) & \( c^\textit{med}_\textit{init} \) & \( c^\textit{med}_\textit{final} \) & \\
\hline
    \( \text{SA}-\mathbb{R}^7 \) & 0.5469 & 20.2001 & 11.9929  & 0.3790 & 23.9119  & 12.3742 & 0.6006 & 22.2275  & 13.3074   & 0.9284  & 26.3252 & 14.8356   & 0.8493  & 24.6474  & 16.3089  \\
    \( \text{DA}-\mathbb{R}^{14} \) & 0.4973  & 19.3292 & 10.2405   & 0.3562 & 23.0785  & 10.6342  & 0.5466 & 22.8266   & 11.7872  & 0.6015  & 23.8139  & 12.5322  & 0.6177  & 24.3534  & \textcolor{red}{12.4140}  \\
   
 \hline
\end{tabular}}
\label{tab:benchmark-Manipulation}
\end{table*}

Fig.~\ref{fig:result-openrave} shows the performance of all algorithms in the Single-Arm Manipulator Problem and in Dual-Arm Manipulator Problem. RRT-Connect and ABIT* are faster than other algorithms but may result in higher costs, making them suitable for applications requiring quick solutions. On the other hand, Lazy PRM*, BIT* and AIT* achieve lower costs but require more computation time, making them ideal for scenarios where path optimality is crucial. RRT*, {RRT}$^\#$, and Informed RRT* do not perform as well in terms of both success rate and path cost. 
As illustrated in Table~\ref{tab:benchmark-Manipulation}, we compare various algorithms for manipulation tasks, focusing on both single-arm (SA) and dual-arm (DA) configurations in high-dimensional spaces. The performance metrics include median initial solution time (\( t^\textit{med}_\textit{init} \)), median initial solution cost (\( c^\textit{med}_\textit{init} \)), and median solution cost over time (\( c^\textit{med}_\textit{final} \)).

In the single-arm scenario (\( \text{SA}-\mathbb{R}^7 \)), {RRT-Connect} demonstrates a relatively low initial solution time of 0.2260s, but its final solution cost remains high at 18.0909. Conversely, {EIT*} offers a more competitive median initial solution time of 0.8493s while achieving a notably lower median solution cost of 16.3089, indicating its efficiency in path optimization.

The dual-arm configuration (\( \text{DA}-\mathbb{R}^{14} \)) presents similar trends. {EIT*} achieves an initial solution time of 0.6177s, which is comparable to other algorithms, while also has the minimal final solution cost at 12.4140. This positions EIT* as a strong contender in both initial speed and long-term cost efficiency.

Among the other algorithms, {RRT*}, RRT$^\#$, {Informed RRT*} exhibit higher initial solution times, indicating less efficiency in quickly generating paths. {LazyPRM*} shows slightly better final solution costs but suffers from higher initial costs and times compared to EIT*.

Overall, these experiments demonstrate the performance of ten popular algorithms across various scenarios, solidifying their status as leading choices in high-dimensional manipulation path planning. The performance in both single-arm and dual-arm tasks highlights their feasibility and efficiency.

\section{Conclusion}
\label{sec:conclusion}


Sampling-based motion planning algorithms are highly effective for exploring continuously-valued spaces, which are commonly encountered in robotics. These algorithms rely on generating samples to approximate and explore the search space. Many sampling-based algorithms are probabilistically complete. But these algorithms do not provide any guarantee on the quality of its solution. In recent years, researchers have focused on addressing this issue. In this article, we have reviewed the progress made. We divide the traditional sampling-based algorithms framework into seven parts, \textbf{\textit{Sampling}}($\cdot$), \textbf{\textit{Nearest}}($\cdot$), \textbf{\textit{Steer}}($\cdot$), \textbf{\textit{CollisionFree}}($\cdot$), \textbf{\textit{Extend}}($\cdot$) including \textbf{\textit{Heuristic}}($\cdot$), \textbf{\textit{Rewire}}($\cdot$). We conduct an extensive literature review on these seven aspects and provide a summary and analysis of the current research status. Furthermore, this paper tests the performance of ten popular planners in both simulated random scenarios and Manipulation tasks with different dimensions. The results suggest that planning algorithms are capable of solving a wide range of problems, including those with narrow passages and constraints. However, no single planner consistently outperforms others across all problem types. Apart from the comparative evaluations, this review provides a comprehensive overview about applications and limitations, such as dynamic environments, Narrow passage and kinodynamic constraints. This helps researchers and practitioners to make better understanding sampling-based planners in different fields. 

In summary, by systematically reviewing the state of the art and addressing the remaining challenges, this survey serves as a valuable resource for researchers and practitioners in the field of robotics. It offers a clear understanding of the current landscape of motion planning techniques and provides insights into the future directions of this rapidly evolving area of study.

\bibliographystyle{IEEEtran}
\bibliography{bibliography}

\end{document}